\documentclass[lettersize,journal]{IEEEtran}
\usepackage{amsmath,amsfonts}
\usepackage{algorithmic}
\usepackage{algorithm}
\usepackage{array}
\usepackage[caption=false,font=normalsize,labelfont=sf,textfont=sf]{subfig}
\usepackage{textcomp}
\usepackage{stfloats}
\usepackage{url}
\usepackage{verbatim}
\usepackage{graphicx} 
\usepackage{threeparttable}
\usepackage{booktabs}
\usepackage{multirow}
\usepackage[inkscapelatex=false]{svg}
\usepackage{enumitem}
\usepackage{xcolor}

\usepackage{hyperref}
\usepackage[style=ieee,minnames=1,maxnames=3,doi=true,eprint=false]{biblatex}
\newcommand{\eq}[1]{Eq.~(\ref{eq:#1})}
\newcommand{\fig}[1]{Fig.~\ref{fig:#1}}
\newcommand{\tab}[1]{Table~\ref{tab:#1}}
\newcommand{\sect}[1]{Section~\ref{sect:#1}}

\newcommand{\etal}{\emph{et al.~}}
\newcommand{\ie}{\emph{i.e.~}}

\newcommand{\egc}{\emph{e.g.,~}}

\usepackage{pifont}
\newcommand{\yes}{\ding{51}}
\newcommand{\no}{\ding{55}}

\newcommand{\tevent}{16}  
\newcommand{\accuracy}{87.8\%}
\newcommand{\prjname}{EvGNN}

\definecolor{matlabBlue}{rgb}{0,0.447,0.741}
\newcommand{\colorblue}{\color{matlabBlue}}
\usepackage{eso-pic} 
\usepackage{textcomp} 

\AddToShipoutPicture*{\small \raisebox{26.7cm}
{\hspace*{2.0cm}
\fboxrule=1.5pt
\fboxsep=0.0pt
\fbox{\colorbox{pink}
{
\begin{minipage}[c]{0.95\textwidth}
\footnotesize
{\setlength{\baselineskip}{0.5\baselineskip}
\vspace*{1pt}
This document is the paper as accepted for publication in \textit{IEEE Trans. Circuits and Systems for AI} (doi:\href{https://doi.org/10.1109/TCASAI.2024.3520905}{10.1109/TCASAI.2024.3520905}).\newline
\textcopyright 2025 IEEE.  Personal use of this material is permitted.  Permission from IEEE must be obtained for all other uses, in any current or future media, including reprinting/republishing this material for advertising or promotional purposes, creating new collective works, for resale or redistribution to servers or lists, or reuse of any copyrighted component of this work in other works.
\par}
\end{minipage}}}}}%

\begin{document}

\title{\vspace*{-2mm}\prjname: An Event-driven Graph Neural Network Accelerator for Edge Vision}

\author{Yufeng Yang$^\dag$, Adrian Kneip$^\dag$,~\IEEEmembership{Member, IEEE}, and Charlotte Frenkel,~\IEEEmembership{Member, IEEE}
\vspace*{-4mm}

\pagenumbering{gobble} 

\thanks{Yufeng Yang was with the Microelectronics Department (EEMCS Faculty), Delft University of Technology, 2628 CD Delft, Netherlands.
Adrian Kneip is with the Microelectronics Department (EEMCS Faculty), Delft University of Technology, 2628 CD Delft, Netherlands, and with the Department of Electrical Engineering, KU Leuven, 3000 Leuven, Belgium.
Charlotte Frenkel is with with the Microelectronics Department (EEMCS Faculty), Delft University of Technology, 2628 CD Delft, Netherlands. (Corresponding author: c.frenkel@tudelft.nl). $^\dag$Co-first authors, contributed equally to this work.}
}

\markboth{}
{Yang \textit{et al.}: Event-driven Graph Neural Network Accelerators for Edge Vision}

\maketitle

\begin{abstract}
 
Edge vision systems combining sensing and embedded processing promise low-latency, decentralized, and energy-efficient solutions that forgo reliance on the cloud. As opposed to conventional frame-based vision sensors, event-based cameras deliver a microsecond-scale temporal resolution with sparse information encoding, thereby outlining new opportunities for edge vision systems. However, mainstream algorithms for frame-based vision, which mostly rely on convolutional neural networks (CNNs), can hardly exploit the advantages of event-based vision as they are typically optimized for dense matrix-vector multiplications. While event-driven graph neural networks (GNNs) have recently emerged as a promising solution for sparse event-based vision, their irregular structure is a challenge that currently hinders the design of efficient hardware accelerators. In this paper, we propose \prjname, the first event-driven GNN accelerator for low-footprint, ultra-low-latency, and high-accuracy edge vision with event-based cameras. It relies on three central ideas: (i)~directed dynamic graphs exploiting single-hop nodes with edge-free storage, (ii)~event queues for the efficient identification of local neighbors within a spatiotemporally decoupled search range, and (iii)~a novel layer-parallel processing scheme allowing for a low-latency execution of multi-layer GNNs. We deployed \prjname~on a Xilinx KV260 Ultrascale+ MPSoC platform and benchmarked it on the N-CARS dataset for car recognition, demonstrating a classification accuracy of \accuracy~and an average latency per event of \tevent$\mathrm{\mu}$s, thereby enabling real-time, microsecond-resolution event-based vision at the edge.

\end{abstract}


\vspace*{3mm}
\section{Introduction}
\vspace*{2mm}
\label{sect:intro}
\IEEEPARstart{E}{dge} vision systems combine digital cameras and embedded computing to capture and process visual information within a limited power budget, typically ranging from milliwatts to a few watts \cite{sparseyolo, tinytracker, pulp, lopecs}. While successful in a wide range of applications, from the Internet-of-Things to robotics \cite{edgeeye, flyingiot, robot_sense_1}, edge vision systems relying on conventional frame-based cameras are ill-suited for latency-critical scenarios requiring decisions within microseconds, such as autonomous driving or drone navigation tasks \cite{navion,arafat_drones_2023,gupta_cars_2021}. Nonetheless, standard frame-based cameras fail to meet these latency requirements as they typically capture 30-60 frames per second (FPS), while the power dissipation of 1000-FPS high-speed cameras can reach tens of watts \cite{1000fps}, far exceeding the power budget of common edge vision systems.

Event-based cameras, also called silicon retinas~\cite{sretina} or dynamic vision sensors (DVSes) \cite{dvs}, have attracted growing interest in low-power high-speed edge vision applications. As opposed to their frame-based counterparts, which record absolute light intensity for every pixel in the field of view at a fixed sampling rate, event-based cameras are locally sensitive: their pixels are individually activated only when the received light intensity changes beyond a preset threshold~\cite{event_survey}. When this threshold is exceeded, event packets containing the pixel address are generated. This event-driven scheme operating at the pixel level leads to (i)~a temporal resolution on the order of microseconds, which helps alleviate motion blur, and (ii)~a power consumption on the order of milliwatts thanks to sparsity~\cite{event_survey}. Indeed, as no event is generated in the absence of light intensity changes, static background information is filtered out: \mbox{only the moving objects are contained in the event stream.}

As the standard neural network model for computer vision tasks, convolutional neural networks (CNNs) have demonstrated excellent performance for scenarios such as object recognition or detection \cite{alex,resnet,yolotiny,mobilenetv3,dualconv,cnn_survey}. However, since they are optimized for the mainstream frame-based vision paradigm, techniques for sparse event-based data are still at an early development stage. Indeed, current event-based vision systems usually rely on a straightforward conversion of event streams into images by accumulating events from each pixel within a given time window, which is also known as the \textit{dense-frame approach} \cite{hugnet_ref_13,time_slice,dense_frame}. While this conversion allows leveraging mature frame-based algorithms to process event-based data, accumulation windows on the order of milliseconds are necessary to obtain reasonable performance \cite{time_slice}, thus discarding the original microsecond-level resolution of the event-based data. 

\begin{figure*}[ht]
    \centering
    \vspace*{3mm}
    \includegraphics[width=0.98\linewidth]{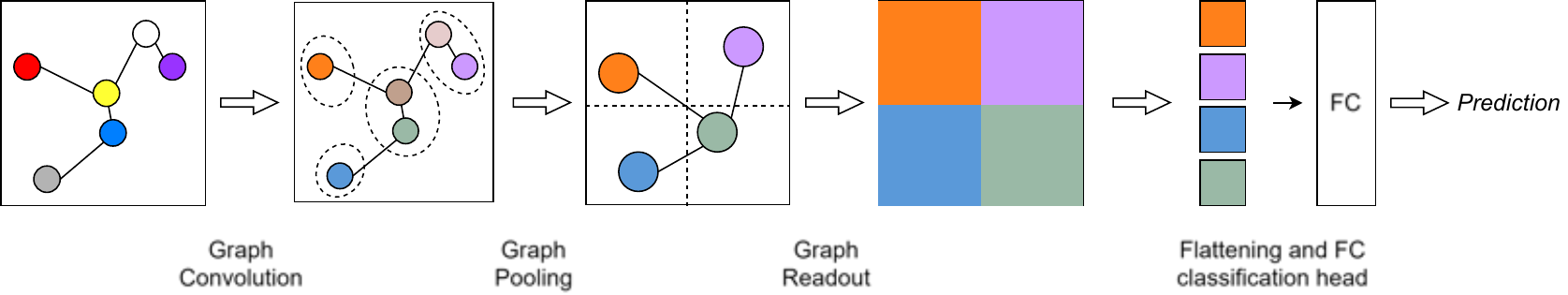}
    \caption{Illustration of a typical event GNN pipeline. The colors of nodes represent their features. After the graph convolution, the features are updated (color changed). The graph pooling layer, where certain nodes are selected and edges are re-arranged, simplifies the graph structure through node down-sampling. The graph readout layer divides the graph into several grids, whose cells are assigned with node features. Finally, after flattening, the cell features are processed by the FC layer to derive the graph-level prediction results.\vspace*{1.5mm}}
    \label{fig:basic_gnn}
\end{figure*}

As solving this challenge requires departing from frame-based vision, graph representations of event-based data have recently started being explored \cite{evs, aegnn, hugnet}. By regarding individual events as \textit{nodes} and their spatiotemporal relationships as \textit{edges}, a stream of events can be represented as a fully equivalent graph (\ie\emph{event graph}) without degrading the temporal resolution nor losing the inherent sparsity of the data. While this sparse event graph is well suited for processing by graph neural networks (GNNs) \cite{gcn, spline}, the standard approach consisting of first constructing the full event graph, and then processing it with a GNN, fails to exploit the event-driven nature of the data and is bound to millisecond-level latencies in dedicated hardware~\cite{flowgnn, blockgnn}. To solve this challenge, techniques to construct and process the event graph in a \emph{dynamic} fashion (\ie on a per-event basis) have recently been proposed~\cite{aegnn, hugnet}, thereby enabling both sparsity-aware and event-driven processing. However, designing efficient hardware accelerators for these sparse, irregular, and fully data-driven workloads, is currently an open challenge.

In this work, we introduce EvGNN, the first \underline{ev}ent-driven \underline{GNN} accelerator for edge vision at microsecond-level latencies that supports the end-to-end hardware acceleration of an event graph, from event-based input acquisition to dynamic event graph construction and real-time GNN inference. Our main contributions are as follows:

\begin{itemize}

    \item We exploit the \textit{causality} of event graphs through directed edges to achieve ultra-low-latency decisions by only processing the local subgraph of direct neighbors around a new event. This preserves accuracy while forgoing the need to store edges, thereby drastically reducing the memory footprint.

    \item We adopt a hardware-friendly \textit{spatiotemporally decoupled prism neighbor search} during the event-graph construction. A dedicated search engine based on cascaded event queues is implemented in hardware to quickly identify valid neighbors in the spatial then temporal dimensions, ensuring the efficient construction of event-wise local subgraphs.
    
    \item We introduce the concept of \textit{layer parallelism} to speed up the processing of event-based GNNs with directed edges, reusing past information on a new event's neighborhood to parallelize the computation of every layer's new features, reducing the end-to-end update latency per event.
    
    \item We finally \textit{deploy} \prjname~on a Xilinx KV260 Ultrascale$+$ MPSoC platform and evaluate our design with the real-world dataset N-CARS \cite{ncars}. We achieve a prediction accuracy of 87.8\% with 8b weights and activations, with an average latency per event of $16\mu$s and a total on-chip memory footprint of 1.76MB. We thereby demonstrate that $\mu$s-level edge vision can be achieved with low-cost hardware solutions, avoiding the need and cost of fabricating a dedicated accelerator on silicon.

\end{itemize}

This paper is organized as follows. Section II introduces key concepts and algorithms of event-based data, event graphs, and GNNs. Our hardware-algorithm co-design approach is described in Section III, which reduces the computational and memory footprints of graph construction and processing steps toward deployment on edge hardware. Section IV covers the proposed \prjname~accelerator, while benchmarking results are finally provided in Section V. Both the proposed algorithm and the hardware are available in open source at {\colorblue{\mbox{\url{https://github.com/cogsys-tudelft/evgnn}}}}.

\section{Background -- Event Graph Construction and Processing Algorithms}
\label{sect:pre_algo}

\pagenumbering{arabic} 
\setcounter{page}{2}

In this section, we first introduce background information for event graphs (Section~\ref{sect:basic_graph}), GNNs (Section~\ref{sect:basic_gnn}), and finally the methods allowing for an event-based construction and processing of event graphs (Section~\ref{sect:event_driven}).

\subsection{Event Graphs}
\label{sect:basic_graph}

A graph $ \mathcal{G=\{V, E\}} $ is a data structure where objects are abstracted as nodes (vertices) $\mathcal{V}$ and their relationships are captured as edges $\mathcal{E}$. Each node in a graph can be assigned additional information, referred to as \emph{features}. Each edge, which connects two nodes, can be either undirected or have a fixed direction pointing from one node to another, thus forming an \emph{undirected} or a \emph{directed} graph, respectively.

Event stream data is a series of events generated from the activated pixels of event-based cameras, in which an event \mbox{$ev = (x,y,t,p)$} is represented as a combination of the pixel's spatial position $(x,y)$, the timestamp $t$, and the binary polarity \mbox{$p = \{0,1\}$} to indicate a positive or negative change in light intensity, respectively.
An event stream can be viewed as a graph, denoted as an \emph{event graph}, where each event is represented as a node with two key features: the spatiotemporal position $(x,y,t)$ and the polarity $p$ \cite{static_event_graph_1, static_event_graph_2}. Nodes connected by an edge are referred to as \emph{neighbors} for each other. For a given node $i$, a subgraph $\mathcal{N}(i)$ consisting of all neighbors and connecting edges is denoted as the \emph{1-hop} \emph{neighborhood} of the node. Similarly, the \emph{2-hop} neighborhood subgraph will include the neighbors' neighbors and their connecting edges, while the \emph{n-hop} neighborhood subgraph will expand to the \mbox{$n^{th}$-order} neighbors of the node~\cite{hop_define}.

\subsection{Graph Neural Networks (GNNs)}
\label{sect:basic_gnn}

For inference on graph data, GNNs gather information from nodes and edges of a graph to generate a new representation of the graph. Similar to CNNs, convolutional GNNs (simply referred to as GNNs hereafter) are composed of graph convolutional layers, graph pooling layers, graph readout layers, and standard fully-connected (FC) layers for the final prediction head (\fig{basic_gnn}):

\subsubsection{Graph convolution}
\label{sect:graphconv}

\begin{figure}[!t]
    \centering
    \hspace*{-0.6cm}  
    \includegraphics[scale=0.55]{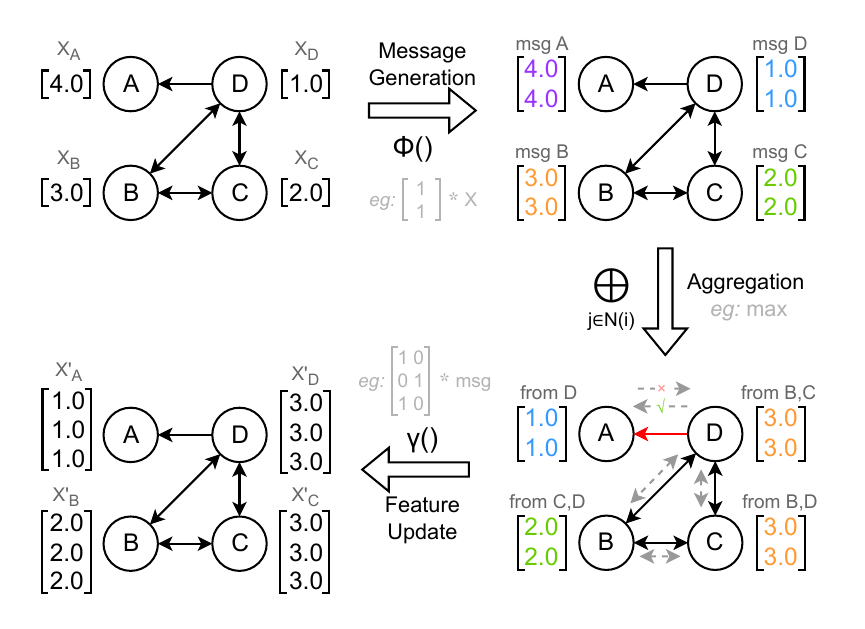}
    \caption{Illustration of graph convolution, consisting of three steps: message generation, aggregation, and feature update. In the graph, each node $i$ has a feature vector. First, every node generates its message (colored numbers) by the differentiable function $\phi()$ (\emph{message generation}, here illustrated with a simple replication operation). Next, messages are exchanged through the edges and nodes aggregate the messages they receive (\emph{aggregation}, illustrated with a max-value operation). Note that the message from $A$ cannot be aggregated by $D$ due to the directed edge. Finally, the aggregated features are transformed by $\gamma()$, generating the new layer's feature vectors (\emph{feature update}, illustrated with a replication operation).}
    \label{fig:message_passing}
\end{figure}

The general description of various graph convolution types, typically expressed as a \emph{message passing algorithm}, consists of three major steps: message generation, aggregation, and feature update~\cite{1stgnn}. As illustrated in \fig{message_passing}, within a graph convolutional layer $l$, these steps are carried out for each node $i$ as per \eq{gnn} \cite{pyg}, ignoring by simplicity any edge feature aside from binary connectivity:

\begin{equation}
\label{eq:gnn}     
\mathbf{x}^{l+1}_i = \gamma \left( \mathbf{x}^l_i,
\bigoplus_{j \in \mathcal{N}(i)} \, \phi
\left(\mathbf{x}^l_i, \mathbf{x}^l_j\right) \right),
\end{equation}

\noindent where $\mathbf{x}^l_i$, $\mathbf{x}^l_j \in \mathbb{R}^{C_{in}}$ are the input feature vectors respectively associated with the current node $i$ and each of its neighbors $j \in \mathcal{N}(i)$, with $C_{in}$ the number of input feature channels, and $\mathbf{x}^{l+1}_i \in \mathbb{R}^{C_{out}}$ is the resulting output feature vector with $C_{out}$ feature channels.

The convolution steps are thus as follows \cite{gnn_survey}:

\begin{enumerate}[label=\roman*)]

    \item \textit{Message generation}: for each neighbor $j$, the features of node $i$ and those of neighbor $j$, ($\mathbf{x}^l_i$, $\mathbf{x}^l_j$), go through a learnable differentiable function $\phi()$ (\egc a linear transformation \cite{gcn} or a multi-layer perceptron (MLP) \cite{pointnet}).

    \item \textit{Aggregation}: as each neighbor $j$ of node $i$ generates one message, all messages are aggregated by a chosen permutation-invariant function $\bigoplus$ (\egc summation, average, or maximum). In the case of directed edges, a message can only be aggregated \mbox{if it points from $j$ to $i$.}

    \item \textit{Feature update}: the aggregated message is transformed by another learnable differentiable function $\gamma()$ to generate the $\mathbf{x}^{l+1}_i$ output feature vector of node $i$ for the next GNN layer.

\end{enumerate}

\vspace*{2mm}

\subsubsection{Graph pooling}
Graph pooling layers down-sample the nodes of the input graph to generate a coarser representation \cite{gnn_survey}. One typical method, cluster-based graph pooling (\egc DiffPool \cite{diffpool} and EigenPool \cite{eigenpool}), maps a group of original nodes into a new node (\emph{cluster}) and gathers their features into the cluster node. The number of nodes decreases, while the connections between original node groups are kept and rebuilt as the edges between new cluster nodes.

\subsubsection{Graph readout}
\label{sect:graphread}

While graphs have a data-dependent number of nodes, the final FC classification head requires inputs with fixed dimensions. To solve this mismatch, graph readout layers transform event graphs into a frame-based representation by globally pooling features of all nodes according to a permutation-invariant function \cite{gnn_survey, agcn, dcnn}. Several variants exist, such as the grid-based graph readout used in the work of Schaefer\textit{~et~al.~}\cite{aegnn}, which operates by first defining a 2D grid based on the spatial position $(x,y)$ of each node, and then by executing the pooling operation for nodes within each cell of the grid.

\subsubsection{FC prediction head}
With the regular feature data provided by the graph readout layer, a set of one or multiple FC layer(s) allows generating the final prediction results.

\begin{figure}[!t]
    \centering
    \hspace*{-0.6cm}  
    \includegraphics[width=0.9\columnwidth]{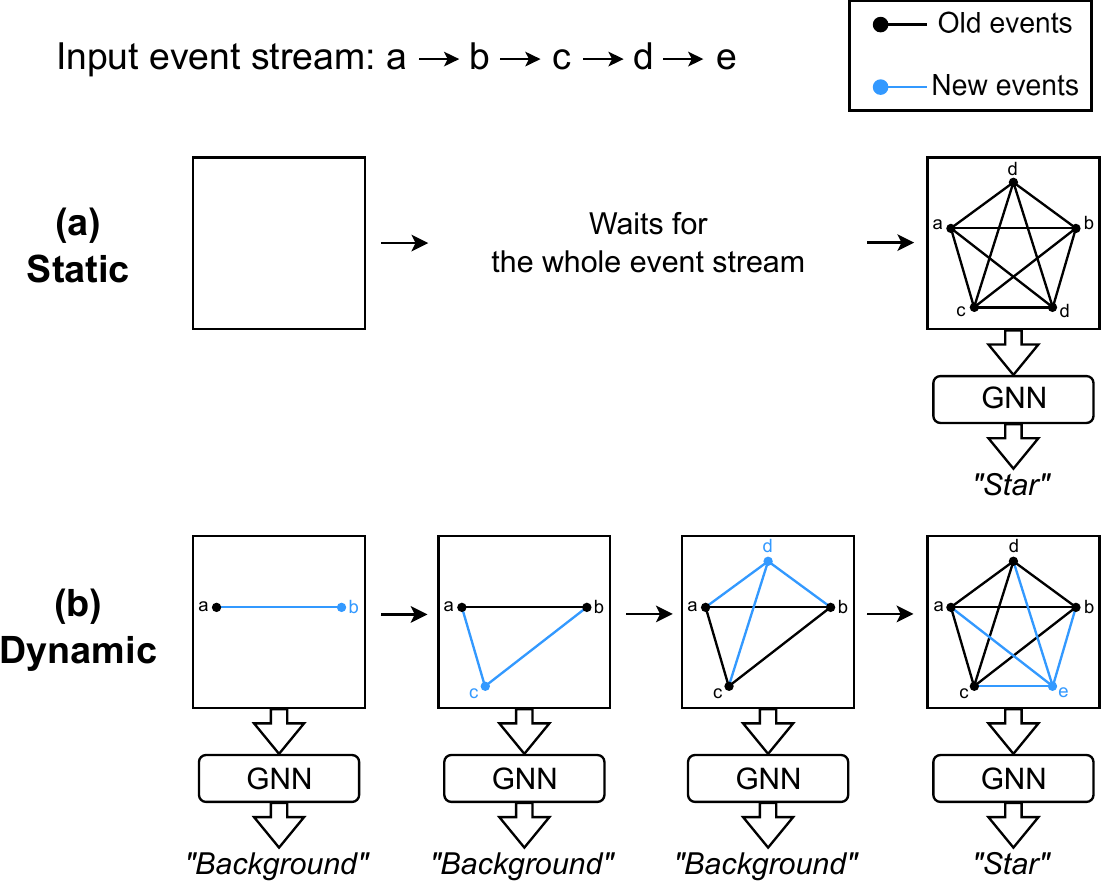}
    \caption{Static and dynamic event graphs generated from the same event stream. (a) The whole event stream is first transformed into a static event graph, then uses a GNN to provide a prediction result. (b) Whenever a new event is generated (shown in blue), the dynamic event graph is updated and then processed by a GNN, thereby generating an updated prediction result on a low-latency, per-event basis.}
    \label{fig:dynamic_event_graph}
\end{figure}

\begin{figure*}[ht]
    \centering
    \includegraphics[width=1.0\linewidth]{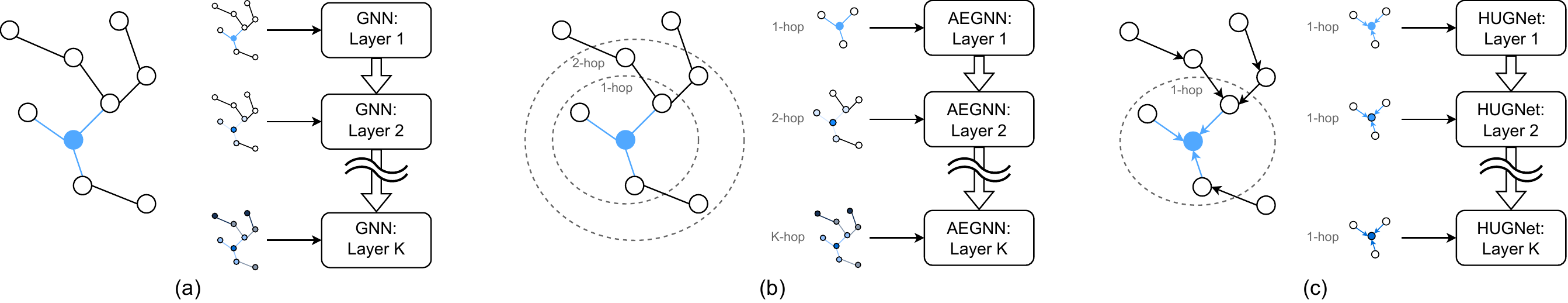}
    \caption{Event-driven K-layer GNN processing schemes. (a)~Naive scheme -- A dynamic event graph, with the blue dot representing the new event node, is processed entirely for each layer of a conventional GNN. The colored nodes represent the updated features along the GNN, while uncolored ones depict unchanged features. (b)~AEGNN scheme~\cite{aegnn} -- The same event graph is processed within a $k$-hop subgraph by the $k^\text{th}$ layer of the event-driven GNN, only involving nodes whose features will be effectively updated. (c)~HUGNet scheme~\cite{hugnet} -- A directed dynamic event graph implies a causal flow of information, hence only the 1-hop subgraph is needed as the input of each layer, because the features of neighboring nodes remain untouched.}
    \label{fig:khop_1hop}
\end{figure*}
\subsection{Event Graph Construction and Processing Strategies}
\label{sect:event_driven}

An event graph first needs to be constructed before it can be processed by a GNN. The \textit{graph construction} (also referred to as \textit{graph building}) step populates an event graph with nodes and connects them with edges if their spatiotemporal distance is within a certain threshold. This step can take place either in a \textit{static} or a \textit{dynamic} fashion~\cite{aegnn,hugnet}. In the former, the entire event sequence is pre-processed and edges are generated according to the spatiotemporal positions of all events. In the latter, the graph is constructed on-the-fly: each new event adds a node and its associated edges to the current dynamic graph. In other words, static graph construction is carried out once the full event stream is available, while dynamic graph construction occurs in an event-driven manner.

The \textit{graph processing} step, based on the execution of a GNN, is strongly dependent on the selected graph construction strategy. Static graph construction is the most common strategy as it allows for the use of vanilla GNNs~\cite{evs, static_event_graph_1, static_event_graph_2}. However, it comes at the expense of latency, as the full event stream needs to be available before a prediction can be generated (\fig{dynamic_event_graph}(a)). To solve this issue, dynamic graph construction has recently been investigated to allow for event-based GNN execution~\cite{aegnn,hugnet}: each new event not only updates the graph, it is also immediately processed by the GNN to update node features and generate a new prediction, a system that we refer to as an \textit{event-driven GNN} (\fig{dynamic_event_graph}(b)).

Naive event-driven approaches, which combine dynamic event graphs directly with conventional GNNs, imply that each layer of the GNN has to process the entire updated graph each time a new event is received (\fig{khop_1hop}(a)), thereby leading to a high computational overhead~\cite{aegnn}. However, as each layer of a GNN only needs information from the 1-hop neighborhood of a node for message passing (Eq. (\ref{eq:gnn})), the final output features of a node in a K-layer GNN are governed exclusively by its K-hop neighborhood (\ie the \textit{receptive field} of the GNN)~\cite{gnn_receptive_1, gnn_receptive_2}. This implies that, for dynamic event graphs, event-driven GNNs only need to process 1-to-K-hop subgraphs for each layer upon the reception of a new event, rather than the whole graph. This technique, introduced in \cite{aegnn} as the asynchronous event-based graph neural network (AEGNN) method and illustrated in \fig{khop_1hop}(b), significantly reduces computation while maintaining mathematical equivalence.

In their hemi-spherical update graph neural network (HUGNet) solution \cite{hugnet}, Dalgaty \etal exploit \emph{directed} dynamic event graphs, where the edges can only point from past nodes to newly added ones (\fig{khop_1hop}(c)). This makes message passing in GNN layers a \textit{causal} process, as opposed to AEGNN where undirected graphs can lead to information being shared from new nodes to past ones, thereby leading to three key advantages. First, HUGNet solves one of the main issues of AEGNN for event graph construction, namely that a new event node needs to wait for future potential neighbors. Second, once a new event is received, only the features of the corresponding new node need to be updated: features of its neighbors within the K-hop range, constituted of past nodes, are fixed as information cannot flow in an anti-causal direction. This implies that the processing range of a K-layer event-driven GNN can be decreased from 1-to-K-hop subgraphs to only the 1-hop subgraph, as illustrated in \fig{khop_1hop}(c). Finally, as the range of message passing is restricted to the 1-hop neighborhood of each new event, edges can be computed on-the-fly for the 1-hop range: they do not need to be stored, nor computed beyond this range.

\section{Hardware-aware graph construction and processing}
\label{sect:codesign}
\begin{figure*}[!t]
    \centering
    \includegraphics[width=0.99\linewidth]{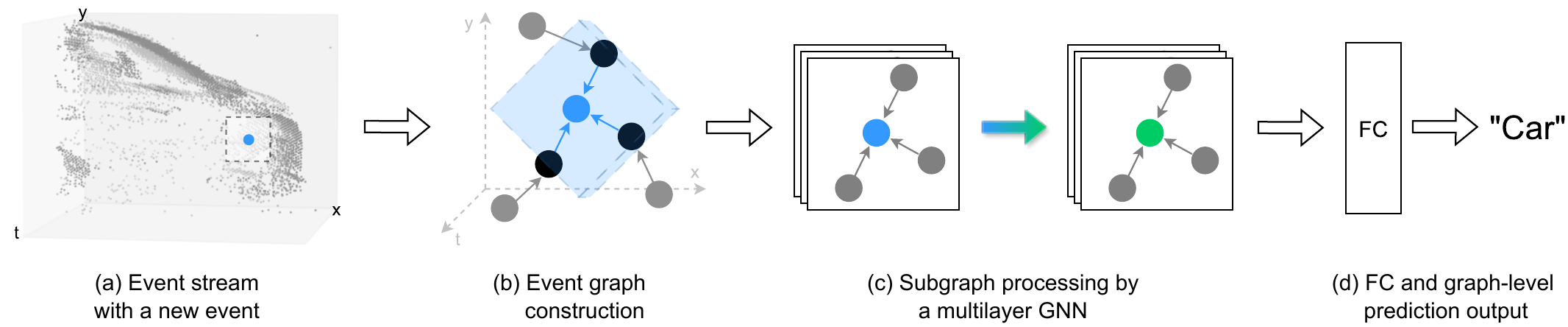}
    \caption{Proposed event-driven GNN processing pipeline. (a)~A new input event (blue) in an event stream, and its neighboring past events, are processed together in an event-driven fashion. (b)~The new event searches potential neighbors in the blue causal prism region, and connects with them by directed edges, thus constructing the event graph (\sect{graph_build_codesign}). (c)~The updated event graph is processed by a multilayer GNN (\sect{network_opt}), where only (i)~the 1-hop subgraph containing the new event is processed, and (ii)~the features of the new node are updated. (d)~The GNN updates the graph-level prediction.}
    \label{fig:event_gnn_general}
\end{figure*}

Previous event-driven GNN algorithms (\egc AEGNN~\cite{aegnn} and HUGNet~\cite{hugnet}) have been developed without a direct consideration for actual hardware limitations faced by a deployment at the edge, thereby making existing algorithms challenging to map on platforms with restricted memory and computational resources. We introduce the first dedicated hardware platform for event-driven GNNs. In this section, we revisit the key steps of graph construction and processing towards an efficient hardware deployment, while minimizing the penalty on accuracy (see~\fig{event_gnn_general} for an illustration of the overall processing pipeline). We select the real-world task of car recognition with event-based vision, benchmarked on the N-CARS dataset~\cite{ncars}. This dataset contains several 100-ms samples of event-based camera recordings, representative of an edge-vision use case. As of today, AEGNN is currently the state-of-the-art event-driven GNN approach on N-CARS~\cite{aegnn}. It will thus be adopted as a baseline, while our co-design developments will be carried out on a validation set from the N-CARS dataset consisting of 15\% of the training set, following a bootstrap strategy~\cite{rinaldo_bootstrap}. All experiments below report the average accuracy obtained over five trials.

\subsection{Graph Construction}
\label{sect:graph_build_codesign}

During the first step of graph construction (Section~\ref{sect:event_driven}), readily executed upon the reception of each new event, spatiotemporal neighbors have to be identified and selected within a certain spatiotemporal distance. This process, known as \textit{radius search}, is however computationally expensive: it usually relies on a \emph{k-d tree} search with frequent insertion and deletion of node data for space partitioning, or on a structural modification of the downstream GNN~\cite{evs}. To solve this challenge, we propose to leverage directed edges and simplified neighborhood search ranges.

\vspace*{1.5mm}
\subsubsection{Directed graph adaptation}
\label{sect:hugnet_adaptation}

\begin{figure}[!t]
    \centering
    \hspace*{-0.6cm}  
    \includegraphics[width=0.95\columnwidth]{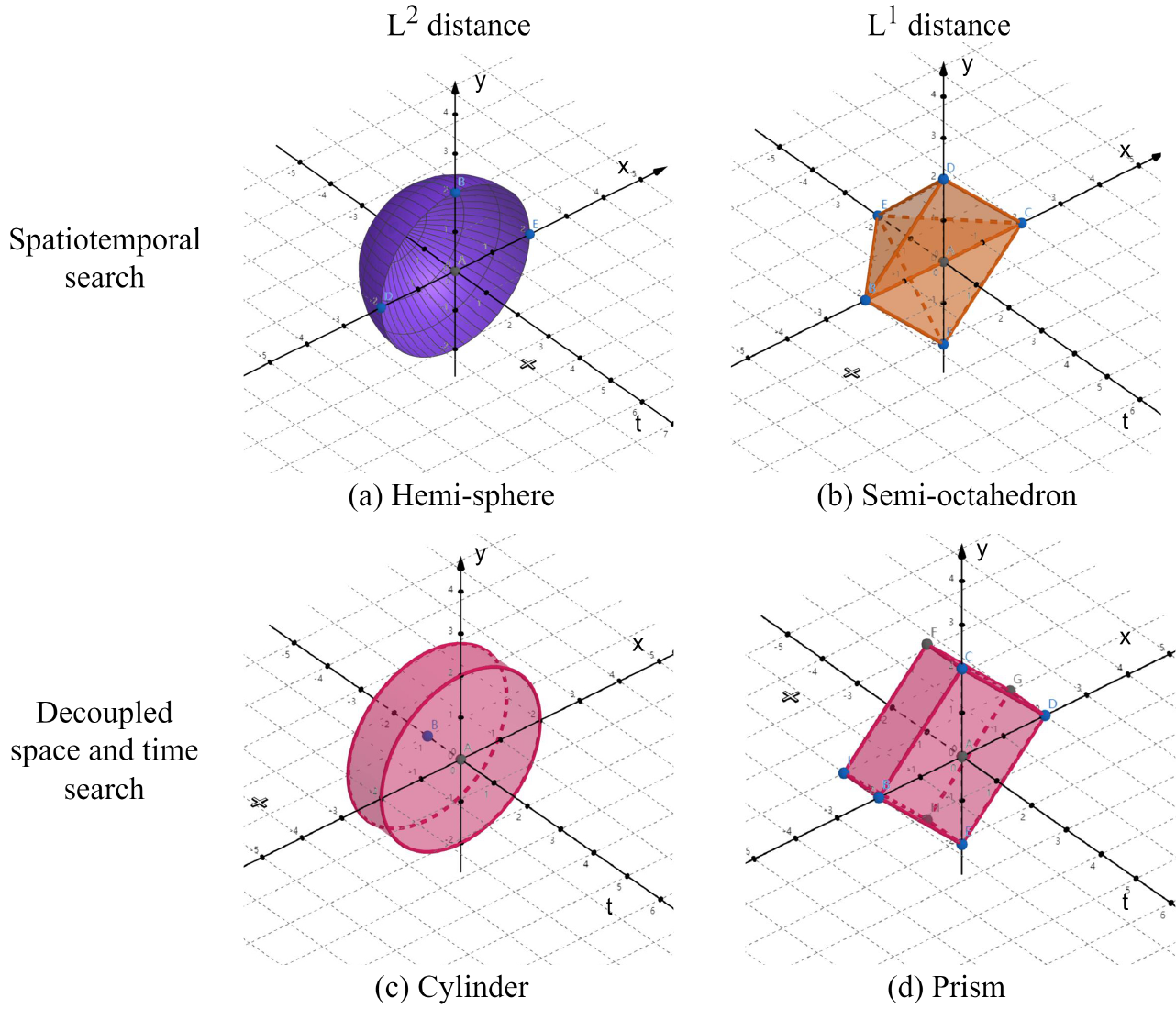}
    \caption{Illustration of four spatiotemporal neighbor search ranges: (a)~hemi-sphere of radius $r$, (b)~semi-octahedron, (c)~cylinder with base radius $r_s$ and height $r_t$, and (d)~prism, obtained by combining two different search schemes with two different $L^p$ distance metrics. Note that the range parameters of all search ranges can be tuned according to the selected application scenario.}
    \label{fig:search_range}
\end{figure}

Inspired by HUGNet~\cite{hugnet}, which was proposed for optical flow estimation tasks (\sect{event_driven}), we adopt the directed graph idea for our classification task scenario. Doing so reduces the spatiotemporal neighborhood search space from a full sphere to a hemisphere (Fig.~\ref{fig:search_range}(a)), as relationships are now causal and no information flows from future nodes to past nodes. Beyond simplifying the search space, we show in \tab{result_directed} that, compared to the AEGNN baseline, adopting directed graphs does not adversely affect accuracy performance.

\begin{table}[!t]
\centering
\caption{Prediction accuracy for AEGNN with directed edges compared to the original AEGNN baseline}
\label{tab:result_directed}
\begin{threeparttable}

\begin{tabular}{cc}
\textbf{Edge type}           & \textbf{Accuracy~$\pm$~Standard Deviation} \\ \toprule
Baseline (Undirected)        & 95.4\% $\pm$ 0.49\%\tnote{*}          \\
Directed                     & 95.7\% $\pm$ 0.26\%           \\  \bottomrule
\end{tabular}

\begin{tablenotes}
    \footnotesize
    \item[*] Based on AEGNN open-source codes on the validation set.
    \end{tablenotes}
\end{threeparttable}
\end{table}

\vspace*{1.5mm}
\subsubsection{Prism neighbor search range}
\label{sect:codesign_search_range}

The \emph{neighbor search range} defines within which spatiotemporal region a previous event can be regarded as a neighbor to a new one. The distance defined by the $L^p$~norm of a difference vector of two position vectors is known as the \emph{$L^p$ distance} \cite{lp}. For nodes in the $\mathcal{R}^3$ spatiotemporal space with position $\mathbf{pos}_i = (x_i, y_i, t_i)$, the raw $L^1$ and $L^2$ distances are%

\begin{equation*}
\begin{split}
   L^1(\mathbf{pos}_1, \mathbf{pos}_2) &= |x_1 - x_2| + |y_1 - y_2| + |t_1 - t_2| \\
   &= |dx| + |dy| + |dt|,
    \\[0.5cm]
    L^2(\mathbf{pos}_1, \mathbf{pos}_2) &= \sqrt{dx^2 + dy^2 + dt^2}.
\end{split}
\end{equation*}

While the $L^2$ distance is prevalent across previous graph-based event vision algorithms \cite{aegnn,hugnet,evs,static_event_graph_1,static_event_graph_2}, it involves the computationally costly square root function to calculate the spatiotemporal neighborhood range of a node. On the contrary, the $L^1$ distance only involves addition and absolute value functions, both of which are hardware-friendly. As illustrated in Figs.~\ref{fig:search_range}(a) and~\ref{fig:search_range}(b) for directed edges, switching from an $L^2$ to an $L^1$ distance transforms the search range from a hemi-sphere to a semi-octahedron.

Complementary to the choice of the distance function, selecting the radius $r$ that bounds the maximum spatiotemporal distance often involves the definition of a proper scaling factor between time and space dimensions, which can have widely different scales (\egc a few pixels vs. thousands of microseconds). Hence, previous works~\cite{aegnn, hugnet} adopted a modified position $\mathbf{pos}^{*}_i$ with a preset time-scaling factor $\beta$ in the $L^2$ distance:

\begin{equation}
\label{eq:normal_dist}
    L^2(\mathbf{pos}^{*}_1, \mathbf{pos}^{*}_2) = \sqrt{dx^2 + dy^2 + (\beta \; dt)^2} \le r.
\end{equation}

\noindent However, given that $\beta$ can compensate for a scale difference of several orders of magnitude, this approach is only practical when high-precision arithmetic is available. Therefore, instead of following a rescaling approach with a single spatiotemporal distance metric of radius $r$, we propose to separate it into two independent parts: the spatial distance range $r_s$ and the temporal distance range $r_t$. In this way, the spatial and temporal metrics can be decoupled, and each of them can be processed in different scales and resolutions. The spatial and temporal conditions for $L^2$ and $L^1$ distance metrics are given in Eqs.~(\ref{eq:l2_cyliner}) and (\ref{eq:l1_cyliner}), and correspond to the cylinder and prism search ranges illustrated in Figs.~\ref{fig:search_range}(c) and~\ref{fig:search_range}(d), respectively. 

\begin{equation}
\label{eq:l2_cyliner}
    \sqrt{dx^2 + dy^2} \le r_s \;\; \mathrm{and} \;\; dt \le r_t 
\end{equation}

\begin{equation}
\label{eq:l1_cyliner}
    |dx| + |dy| \le r_s \;\; \mathrm{and} \;\; dt \le r_t 
\end{equation}

\begin{table}[!t]
\centering
\caption{Prediction accuracy for directed-edge AEGNN with different neighbor search ranges}
\label{tab:search_range}
\begin{tabular}{cccc}
\textbf{Search Ranges} & \textbf{Search Scheme}       & \textbf{$L^p$} & \textbf{Acc.~$\pm$~S.D.} \\ \toprule
Hemi-sphere            & \multirow{2}{*}{Spatiotemporal} & $L^2$          & 95.7\% $\pm$ 0.26\%            \\
Semi-octahedron        &                              & $L^1$          & 95.9\% $\pm$ 0.39\%            \\
Cylinder               & \multirow{2}{*}{Spatiotemporally-decoupled}    & $L^2$          & 95.5\% $\pm$ 0.41\%            \\
Prism                  &                              & $L^1$          & 95.1\% $\pm$ 0.45\%            \\ \bottomrule
\end{tabular}
\end{table}
\tab{search_range} provides the prediction accuracies of all four search ranges summarized in \fig{search_range}. The proposed prism search range offers a lower footprint for deployment on custom hardware with an accuracy drop limited to 0.6\%.

\subsection{GNN Architecture Search and Optimization}
\label{sect:network_opt}

Once the graph construction phase for a new event is over, graph features can be extracted by executing the GNN algorithm. Two key design choices influencing the tradeoff between performance and resource usage in GNNs are (i)~the type of graph convolution operation, and (ii)~the overall network architecture, which we investigate hereafter.

\vspace*{1.5mm}
\subsubsection{Graph Convolution}
\label{sect:codesign_graphconv}
A standard baseline for graph convolutions is the one proposed for the \textit{graph convolution network} (GCN)~\cite{gcn}. Based on Eq.~(\ref{eq:gnn}), its node feature update at layer $l$ can be expressed for unitary edge and linear operators~as

\begin{equation}
\label{eq:gcn}
    \mathbf{x}^{l+1}_i = \mathbf{\Theta}^\mathbf{T}_l \cdot \left( \sum_{j \in
    \mathcal{N}(i) \cup \{ i \}} \frac{\mathbf{x}^l_j}{\sqrt{(d_j+1)(d_i+1)}} \right),
\end{equation}

\noindent where $d_i, d_j$ are the \textit{degrees} (\ie the numbers of neighbors) of nodes $i$ and $j$, respectively, serving to normalize the features of each neighbor $j$ (\ie message generation), before summing them (\ie aggregation) and applying a linear transformation with learnable parameters $\mathbf{\Theta}_l \in \mathbb{R}^{C_{in} \times C_{out}}$ (\ie feature update). This operation is applied for each convolutional layer~$l$ of the GCN. While Eq.~(\ref{eq:gcn}) only depends on the neighbors' features, and thus hardly exploits information related to the relative positions between two nodes, recent event-based GNNs such as AEGNN \cite{aegnn,hugnet} have aimed to explicitly exploit this information via \textit{SplineConv}, a graph convolution method derived from \cite{spline} that encodes the positional information of neighboring nodes using B-spline kernels, on top of the nodes' features. The wider representation ability unlocked by this encoding was shown to help improve the classificaton accuracy of event-based GNNs on several tasks, as highlighted in Table~\ref{tab:graphconv}. Nonetheless, B-splines bring a significant computational burden as well as a vast parameter overhead, being hardly compatible with low-latency and low-footprint event-based vision at the edge.

\begin{table}[!t]
\centering
\caption{Performance metrics of different types of graph convolution}
\label{tab:graphconv}
\begin{tabular}{cccc}

\textbf{Convolution Types}  & \textbf{Acc.~$\pm$~S.D.} & \textbf{\#Parameters} \\ \toprule
Baseline (SplineConv) & 95.4\% $\pm$ 0.49\%       & 30.4k           \\
GCN conv        & 92.5\% $\pm$ 1.09\%           & 4.7k            \\
This work  & 95.1\% $\pm$ 0.43\%           & 4.8k        \\ \bottomrule
\end{tabular}
\end{table}

Alternatively, Qi \textit{et al.} introduced \textit{PointNetConv} \cite{pointnet}, a variant of graph convolution supporting positional encoding. In each of the GCN layers, PointNetConv convolutions concatenate the relative position of each neighbor together with its corresponding features, before applying a learnable transform and a max-value aggregation. The updated features of node $i$ for layer $l+1$ are thus obtained as

 \begin{equation}
 \label{eq:pointnetconv}
         \mathbf{x}^{l+1}_i = \gamma_{\mathbf{\Theta}_0}^l \left( \max_{j \in
         \mathcal{N}(i) \cup \{ i \}} \phi_{\mathbf{\Theta}_1}^l ( \mathbf{x}^l_j,
         \mathbf{pos}_j - \mathbf{pos}_i) \right),
 \end{equation}

\noindent where $\gamma_{\mathbf{\Theta_0}}^l, \phi_{\mathbf{\Theta_1}}^l$ are multi-layer perceptrons (MLPs) with weights $\mathbf{\Theta}_0$ and $\mathbf{\Theta}_1$, respectively. Such a graph convolution avoids the resource-intensive B-splines while properly beholding spatial information. Still, it requires to store large end-to-end MLP models. Therefore, we introduce here a simplified version of Eq. (\ref{eq:pointnetconv}), in which $\phi$ is reduced to a single linear layer and $\gamma$ performs the identity operation. Hence, the features update becomes

\begin{equation}
\label{eq:simple_pointnet}
        \mathbf{x}^{l+1}_i =  \max_{j \in
        \mathcal{N}(i)} \Big( \mathbf{\Theta}^\mathbf{T}_l \cdot \big( \mathbf{x}^l_j \, , \,
        |dx_{i,j}|, |dy_{i,j}| \big) \Big).
\end{equation}

Table \ref{tab:graphconv} shows that, while using 6.3$\times$ less parameters than the SplineConv-based AEGNN baseline, our simplified PointNetConv in Eq. (\ref{eq:simple_pointnet}) still achieves a comparable accuracy. This further underlines the importance of preserving spatial information along the GCN layers.

\begin{figure}[!t]
    \centering
    \hspace*{-0.6cm}  
    \includegraphics[width=0.8\columnwidth]{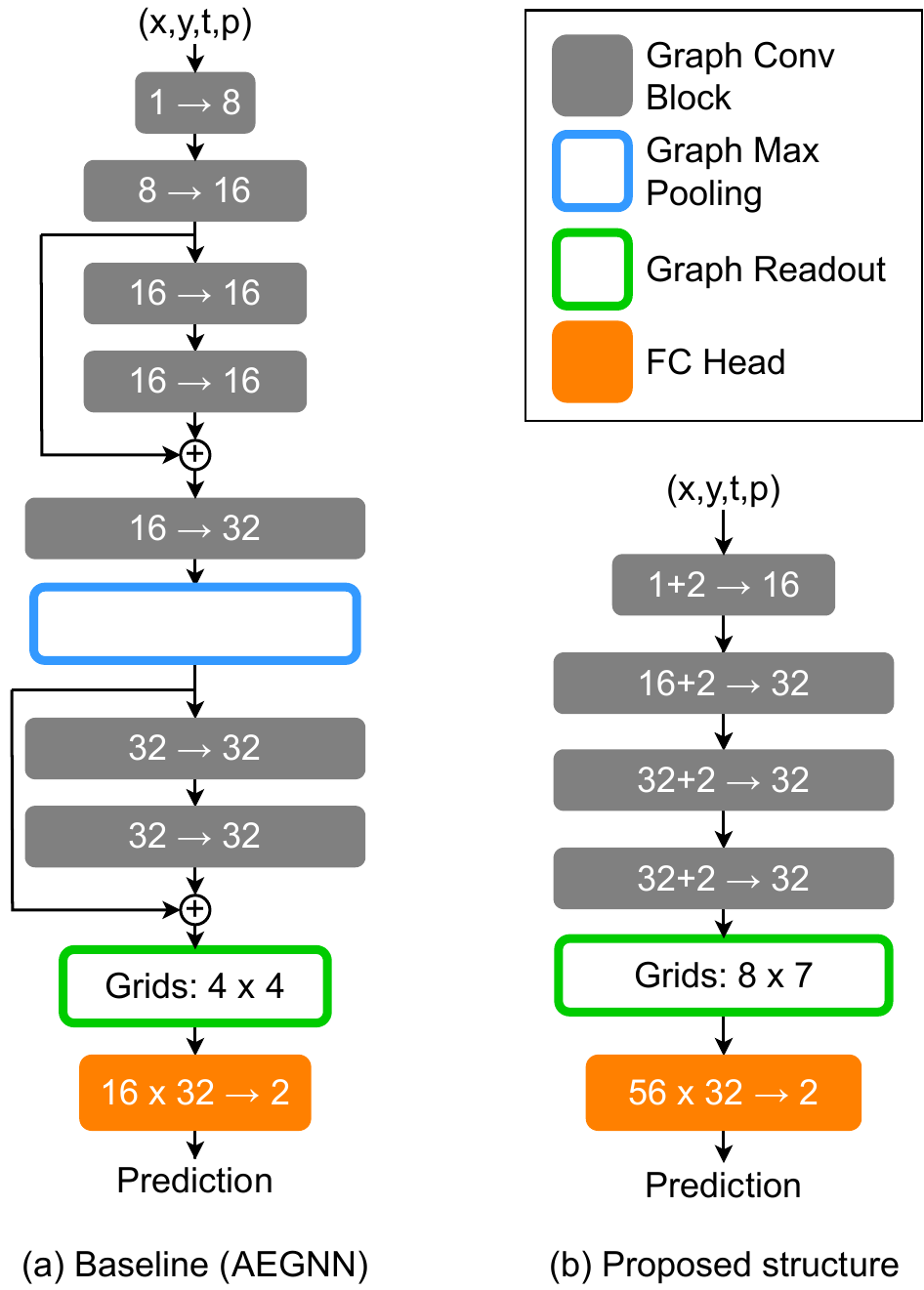}
    \caption{GNN architecture of (a)~the baseline (AEGNN), and (b)~the proposed simplified network. A \emph{Graph Conv Block} contains a graph convolution, an activation function, and a batch normalization layer, where batchnorm folding~\cite{quant3} can be applied. Numbers labeled on graph convolution blocks indicate the corresponding \textit{input} → \textit{output} channel dimensions. The +2 part in (b) corresponds to the concatenated position differences in \eq{simple_pointnet}.}
    \label{fig:my_network}
\end{figure}

\begin{table}[!t]
\centering
\caption{Performance metrics for different network structure simplifications}
\label{tab:net_simp}
\begin{tabular}{ccc}
\textbf{Structure} & \textbf{Acc.~$\pm$~S.D.}  & \textbf{Param. Mem. Footprint} \\ \toprule
Baseline (AEGNN, FP32)   & 95.4\% $\pm$ 0.49\%            & 121.6kB  \\
This work (FP32)        & 95.5\% $\pm$ 0.50\%            & 26.4kB   \\ 
This work (INT8)        & 95.4\% $\pm$ 0.35\%            & 6.6kB   \\  \bottomrule
\end{tabular}
\end{table}

\pagebreak

\subsubsection{Network Architecture}
\label{sect:codesign_network_structure}

Toward a low-cost hardware implementation, we simplify the AEGNN architecture as presented in \fig{my_network}. We show that (i)~using a shallower structure based on PointNetConv graph convolution layers, (ii)~removing residual connections and graph pooling layers, and (iii)~applying batchnorm folding \cite{quant3} together with post-training quantization \cite{quant1} to INT8, allows reducing the required number of parameters by $5\times$ without adversely affecting the classification accuracy (\tab{net_simp}). Note that, since quantizing from 32-bit floating-point (FP32) to INT8 only induces a \mbox{0.1-\%} accuracy drop, FP8 has not been considered an option worth investigating. Indeed, FP8 MAC units require 23$\times$ more resources than their INT8 counterpart when targeting an FPGA implementation \cite{deep_positron} and suffer from a 3$\times$ higher delay per MAC operation \cite{int8_edp}, which would adversely impact latency for negligible accuracy gains over INT8.

\section{Proposed Hardware Architecture for Event-driven GNNs}
\label{sect:hw}

In this section, we design a hardware architecture that accelerates the graph construction (\ie building) and processing algorithms introduced in Section III for a low-power, low-latency deployment at the edge. The proposed hardware accelerator, EvGNN, targets modern Xilinx MPSoC platforms, serving here as demonstrators for $\mu$s-level latency in low-cost edge vision.

\vspace*{1.5mm}
\subsection{Overall System Architecture}
\label{sect:hw_overall}

\begin{figure}[!t]
    \centering
    \hspace*{-0.6cm}  
    \includegraphics[width=0.95\columnwidth]{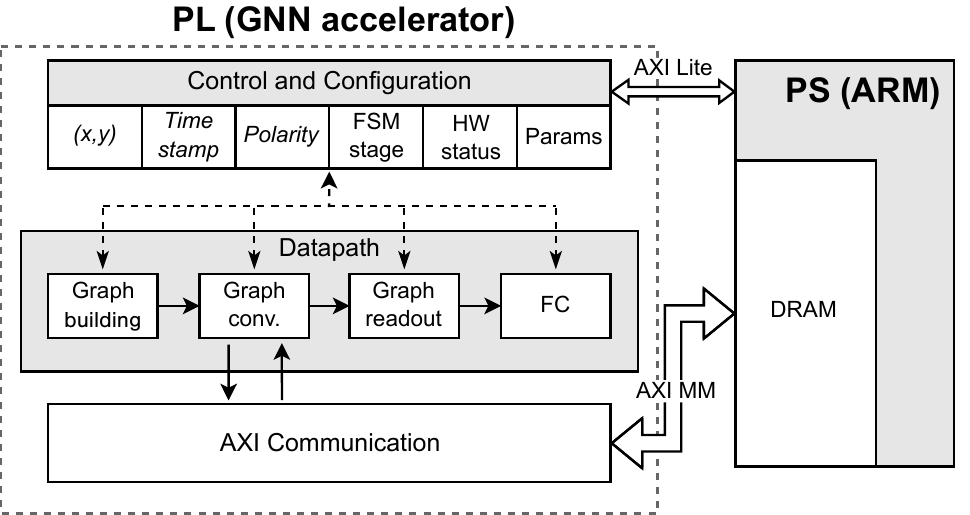}
    \caption{Overall system architecture and partition on the target host FPGA. The \prjname~accelerator is located in the PL, while the PS host CPU is mainly responsible for benchmarking and monitoring.}
    \label{fig:overall}
\end{figure}

Xilinx MPSoC platforms contain \textit{Programmable Logic (PL)} together with a \textit{Processing System (PS)}. A system overview is shown in \fig{overall}: our EvGNN hardware accelerator is located in the PL, while an ARM core in the PS acts as a host CPU that is responsible for (i)~loading dataset samples from the external DRAM, (ii)~feeding data to and reading prediction results from EvGNN through AXI system buses, and (iii)~calculating the prediction accuracy and measuring the runtime. 

The EvGNN accelerator consists of three major blocks. The \emph{Datapath} block is the core implementation of our event-driven GNN algorithm in hardware. It mainly consists of a graph building module, a graph convolution module, as well as graph-readout and FC modules for classification. The \emph{Control and Configuration} block receives event-based data and configurable parameters from the PS, broadcasts them to the \textit{Datapath} block, then sends the prediction results back to the PS. Finally, the \emph{AXI Communication} block implements AXI protocols, including data buffers. In the following, we will elaborate on the modules of the \textit{Datapath} block for graph building and GNN computation.

\begin{figure}[!t]
    \centering
    \includegraphics[width=0.95\columnwidth]{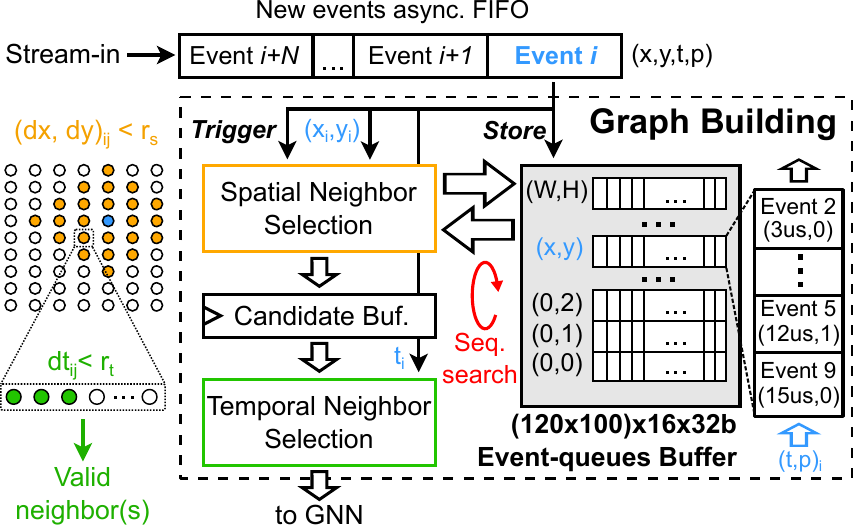}
    \caption{Hardware diagram of the graph-building module, highlighting the execution flow. For each new event $i$, a two-step neighbor-search process is carried out among past events, stored in an on-chip buffer following an event-queue format (W$\times$H, i.e. 120$\times$100 queues of 16 entries each in the N-CARS case, to match the DVS camera's dimensions). The search for neighbors follows the spatiotemporally decoupled prism shape, using the event's coordinates $(x_i, y_i, t_i)$. The coordinates of the valid neighbors are buffered and sent to the graph convolution unit for subsequent processing.}
    \label{fig:hw_graphbuild_diagram}
\end{figure}

\vspace*{1.5mm}
\subsection{Graph-building Module}
\label{sect:hw_graphbuild}

As introduced in Section~\ref{sect:graph_build_codesign}, the first step is to build a local, directed dynamic event graph upon the reception of each new event. A block diagram of the graph-building module is shown in \fig{hw_graphbuild_diagram}, together with an illustration of its execution flow. This module is based on two core design decisions regarding (i)~the storage of the event graph, and (ii)~the selection of neighbors for each new event.

\vspace*{1.5mm}
\subsubsection{Event queues for graph storage}

As the use of directed graphs allows us to forgo the storage of edges (\sect{event_driven}), only the nodes need to be stored and accessed, which can be done in a straightforward manner with \emph{event queues}~\cite{evs, eventq}, as depicted in \fig{hw_graphbuild_diagram}. All queues are stored in a \emph{global event-queues buffer}, where each queue is associated with a pixel of the input DVS camera and stores events whose spatial position is the same as this pixel's physical position $(x,y)$. Event information other than the spatial coordinates is stored in the queue entries, including the timestamp $t$, the polarity $p$, and the event index in the event stream $n$, numbered in chronological order. Pushing a new event pops the oldest one if the queue is full, whereas every entry can be independently read-accessed.

\begin{figure*}[!t]
    \centering
    \includegraphics[width=0.99\linewidth]{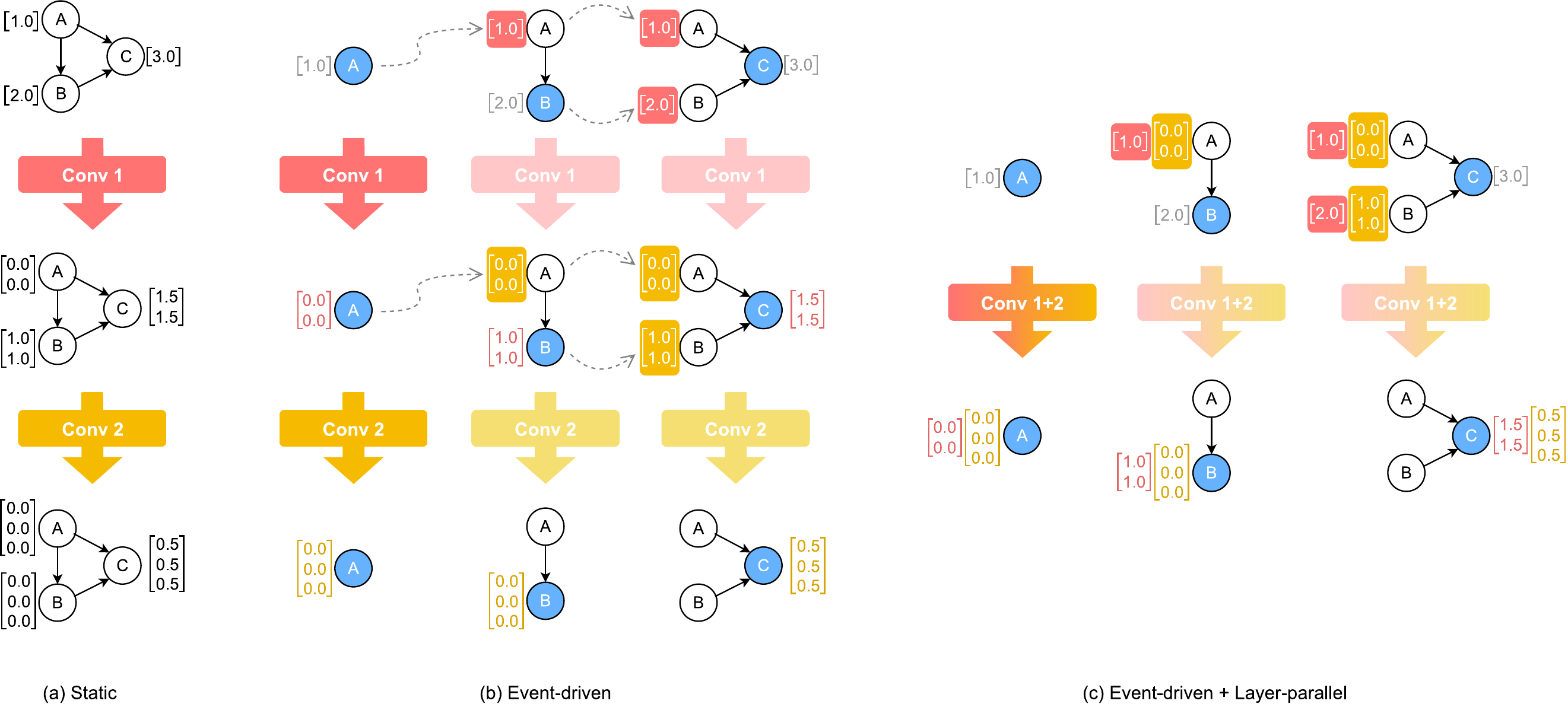}
    \caption{Event graphs processed by an example GNN containing two graph convolutional layers, \emph{Conv 1} and \emph{Conv 2}, both of which consist of a simple average-and-replicate operation for illustration purposes. (a)~A directed \textit{static} event graph is processed by the GNN. Each graph convolutional layer is applied sequentially, while computation of new features can be parallelized over nodes. (b)~A directed \textit{dynamic} event graph is processed by the GNN in an event-driven fashion, with new events colored in blue. The inputs of layers have colored backgrounds with white fonts, while the outputs of layers have white backgrounds with colored fonts. The outputs are identical to those computed with the static graph. Note that the output of a convolutional layer is only used as an input by the \textit{next} layer for \textit{future} events, as shown with dotted gray arrows. (c)~The proposed layer-parallel execution of the same directed graph processed by an event-driven GNN, which is mathematically equivalent to (b).}
    \label{fig:layer_para}
\end{figure*}

\vspace*{1.5mm}
\subsubsection{Spatiotemporally decoupled neighbor selection}

\begin{figure*}[!t]
    \centering
    \includegraphics[width=0.95\linewidth]{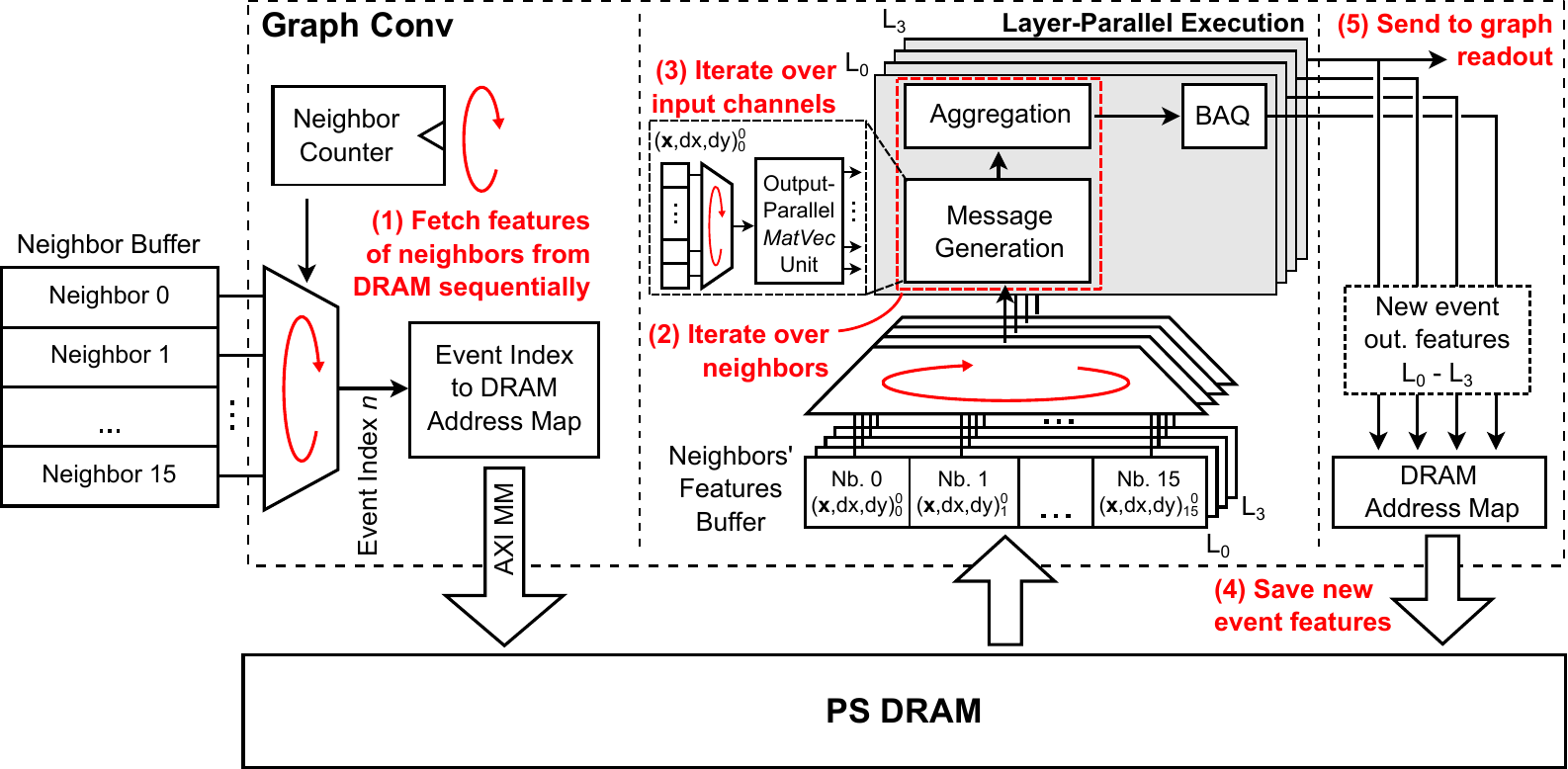}
    \caption{Hardware diagram of the graph convolution module, corresponding to the four-layer network architecture in \fig{my_network}. Node features are first loaded from off-chip DRAM to the graph convolution module through AXI MM buses (i), according to the nodes in the neighbor buffer. Afterwards, for every neighbor (ii), the graph convolution engine performs message passing, iterating over every input channel (iii) before performing the message aggregation. The execution of these steps is parallelized in the \textit{MatVec} unit across the output channel dimension as well as every layer, according to our layer-parallel execution scheme. Then, bias-activation-quantization (\emph{BAQ}) is applied to generate features of the new event, which are stored back into the DRAM (iv). The last layer's updated feature is eventually sent to the readout unit (v).}
    \label{fig:hw_graphconv_diagram}
\end{figure*}

In order to implement the spatiotemporally decoupled prism search range introduced in \sect{graph_build_codesign}, we leverage the fact that the event queue storage scheme intrinsically decouples space and time: spatial information is found in the event queue index, while temporal information is found inside the entries of a queue. This allows the graph-building module to perform the spatial and temporal search steps of \eq{l1_cyliner} independently. First, according to the $(x_i,y_i)$ location of the new event, \textit{spatial search} (\fig{hw_graphbuild_diagram}, orange) is performed to pick all event queues within an $L^1$ spatial distance $r_s$ from the new event. They are accessed from the event-queues buffer and transferred to a \textit{candidate-events buffer}, skipping out-of-bound locations around the target node. Then, the \emph{temporal search} process (\fig{hw_graphbuild_diagram}, green), which can be pipelined with the spatial one, subtracts the timestamps $t_j$ of each candidate event from the current timestamp $t_i$. The obtained time difference $dt_{ij}$ is then compared to the temporal radius $r_t$ to select valid neighbors amongst candidates, which are sent to the graph-convolution module and stored in the graph convolution unit's neighbor buffer (Fig. \ref{fig:hw_graphconv_diagram}, left). The buffer's depth $D_{max} = 16$ bounds the maximum number of neighbors per new event, triggering an early stopping of the search process once full.

\vspace*{1.5mm}
\subsection{Graph-convolution Module}

In the graph-convolution module, we first introduce the concept of \emph{layer-parallel execution} in event-driven GNNs. Then, we propose a hardware architecture and computation flow for this module that exploits this layer-wise parallelism.

\vspace*{1.5mm}
\subsubsection{Layer-parallel execution}

When using GNNs with static directed graphs (\fig{layer_para}(a)), the features of all nodes change after each convolutional layer. As updating the node features of the current layer depends on the new features of the previous one, layers need to be executed sequentially. However, when using dynamic directed event graphs, the causal nature of the graph prevents any feature update of the previous nodes, such that only the features of new incoming events have to be processed, based on their local neighborhood (\fig{layer_para}(b)). Moreover, for convolution operations such as Eq.~(\ref{eq:simple_pointnet}), the computation of output features related to a new event only depends on the node's past neighboring node features, thereby removing any data dependency between GNN layers to compute a new node's features. Hence, as shown in \fig{layer_para}(c), each layer's features associated with the new node's neighbors can be first retrieved, before executing all graph convolutional layers in parallel to get their new node's features. This layer-parallel approach bounds the total execution time to that of the largest GNN layer, and thus drastically shortens the overall per-event latency in multi-layer GNNs while leaving the total number of operations unchanged.

\vspace*{1.5mm}
\subsubsection{Graph-convolution hardware architecture}

The hardware diagram of our graph-convolution module with layer-parallel execution is shown in \fig{hw_graphconv_diagram}. It involves three main stages, which can be pipelined.

\begin{figure}[!t]
    \centering
    \includegraphics[width=0.95\columnwidth]{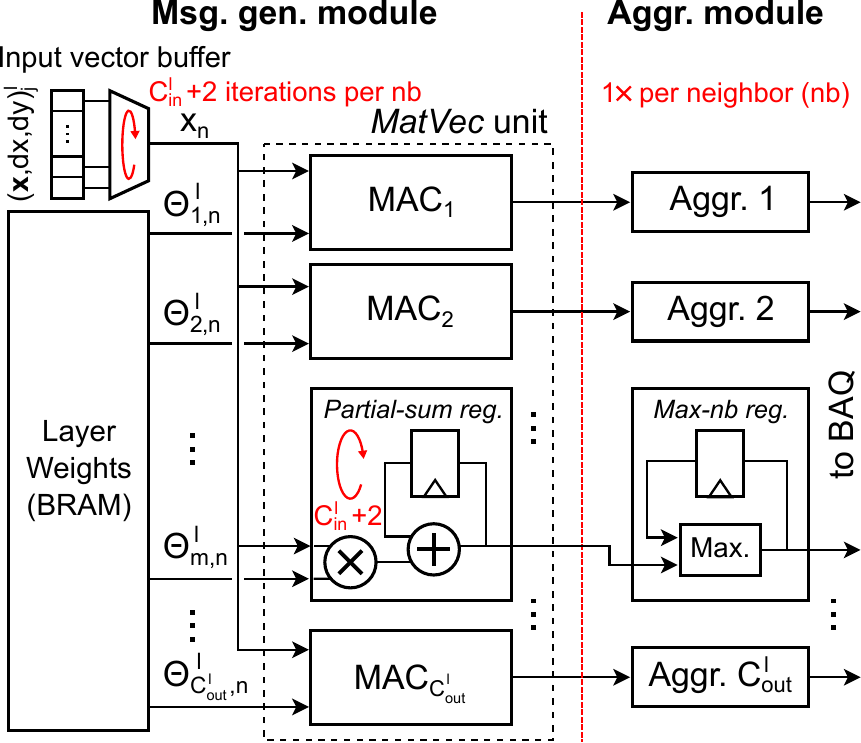}
    \caption{Hardware diagram of the message generation (Msg. gen.) and aggregation (Aggr.) sub-units, including a \textit{MatVec} unit and BRAM weight storage. The input vector ($\mathbf{x}_j, dx_{ij}, dy_{ij}$) of the selected neighbor $j$ contains the $C_{in}$ concatenated node features of the current layer $l$ and the node's position difference (see \eq{simple_pointnet}). The \textit{MatVec} unit performs $C_{in}+2$ sequential accumulations, parallelized across the output feature dimension using $C_{out}$ MAC units. The $C_{out}$ output messages are then broadcast to aggregation units, which store for each output channel the max-valued message amongst the sequentially-addressed neighbors.}
    \label{fig:hw_matvec_diagram}
\end{figure}

\begin{enumerate}[label=\roman*)]
    \item \textit{Neighbors features retrieval --} Neighbors are sequentially popped from the neighbors buffer, one at a time (1). The features of the selected neighbor, previously computed for all graph convolutional layers, are fetched from the external DRAM, which is accessed through an AXI MM bus via the PS.

    \item \textit{Layer-Parallel Execution --} This step implements the graph convolution of \eq{simple_pointnet}, parallelized across all four GNN layers as per the selected network architecture (\fig{my_network}). The features vector $(\mathbf{x}_j, |dx_{ij}|,|dy_{ij}|)^l$ of each neighbor, including its relative position information, are sequentially broadcast to the message generation and aggregation sub-units for every layer $l$ in parallel (2). As detailed in \fig{hw_matvec_diagram}, each message-generation unit performs a matrix-vector multiplication between a neighbor's feature vector and the weight matrix $\mathbf{\Theta}^l$ of that layer, stored in a local BRAM. This operation takes place in $C_{in}^l+2$ cycles and is parallelized across all $C_{out}^l$ output channels by means of a \textit{MatVec} unit, which embeds one digital multiply-and-accumulate (MAC) engine per output channel (3). Then, the aggregation unit receives the newly generated $C_{out}$ messages and performs a max-valued comparison, so as to sequentially keep track of the maximum element per output channel across all neighbors (2). Finally, a \textit{bias} term, ReLU \textit{activation} function, and \textit{quantization} to INT8 is applied to each selected output, which are abbreviated as \textit{BAQ} in \fig{hw_graphconv_diagram}.

    \item \textit{New event features storage --} The outputs of the four \textit{BAQ} units are the new event's output feature vectors, one for each layer. These features are eventually sent back to the external DRAM over the AXI MM bus (4), while the last layer's output is sent to the graph-readout module to serve in the graph prediction update (5).

\end{enumerate}

\vspace*{1.5mm}
\subsection{Graph-readout and FC Modules}

Once a new event has been received and processed by the \textit{Graph-building} and the \textit{Graph-convolution} modules, the prediction of the overall graph can be updated, which is carried out by the \textit{Graph-readout} and \textit{FC} modules (\fig{overall}). The former implements a grid-based graph-readout layer (\sect{graphread}), which divides the full 120$\times$100-pixel input range into an 8$\times$7 grid and selects the maximal features within 16$\times$16-pixel patches. The latter then implements the FC prediction head that provides classification results. In order to carry out the matrix-vector multiplication between the FC weight matrix and the output of the readout layer, we reuse the MatVec unit presented in \fig{hw_matvec_diagram}.

\section{Results}
\label{sect:results}
We evaluate the performance of the designed EvGNN accelerator on the N-CARS dataset after implementing it on a resource-constrained edge platform. Hereafter, we first introduce the experimental setup before assessing the achieved performance and comparing it to the state of the art.

\subsection{Experimental Setup}
\label{sect:exp_setup}

On the software side, experiments are carried out on NVIDIA RTX A6000 GPUs while implemented, trained, and tested using the PyTorch Geometric (PyG) library \cite{pyg}. Similarly to AEGNN~\cite{aegnn}, we first train our event-driven GNN using static event graphs pre-compiled from the N-CARS event streams, where the total \mbox{N-CARS} training data (15,422 event stream samples) is divided into $85\%$ as a training set and $15\%$ as a validation set, following the bootstrapping approach (Section~\ref{sect:codesign}). We use a batch size of 64 and an initial learning rate of 0.002 on the Adam optimizer with 100 epochs. After training, we benchmark our event-driven GNN for inference using the N-CARS test set (8,607 event stream samples), where we represent data as dynamic event graphs.

On the hardware side, the \prjname~accelerator is deployed on a Xilinx KV260 development board, which contains a Zynq UltraScale+ MPSoC in a Kria K26 System-On-Module platform, targeting edge vision applications. As outlined in \sect{hw}, \prjname~is implemented in the PL while an ARM core in the PS takes care of feeding data to EvGNN and collecting its performance metrics, such as the runtime per sample and the overall classification accuracy.

\subsection{Implementation and Benchmarking Results}
\label{sect:bench}

\begin{table}[!t]
\centering
\caption{Resource usage of \prjname~on the Xilinx KV260 platform.}
\label{tab:usage}

\begin{tabular}{ccccc}
\multicolumn{2}{c}{\textbf{Resources}}                                               & \textbf{Used} & \textbf{Available} & \textbf{Utilization} \\ 
\toprule
\multirow{3}{*}{Logic} & LUT      & 30,908  & 117,120 & 26.4\% \\
                        & FF       & 24,083  & 234,240 & 10.3\% \\
                        & DSP      & 228     & 1,248   & 18.3\% \\ 
\midrule
\multirow{3}{*}{\begin{tabular}[c]{@{}c@{}}On-chip\\ memory\end{tabular}} & LUTRAM & 0.45 kB        & 0.44 MB        & 0.1\%               \\
                        & BRAM     & 85.2 kB & 0.63 MB & 13.2\% \\
                        & UltraRAM & 1.68 MB & 2.25 MB & 75.0\% \\ \bottomrule
\end{tabular}

\end{table}

\begin{table}[!t]
\centering
\caption{Summary and performance comparison of state-of-the-art event-driven approaches on the N-CARS test set.}
\label{tab:accuracy}
\begin{threeparttable}
\resizebox{1.0\columnwidth}{!}{
\begin{tabular}{lcccc}
\textbf{Networks} & \textbf{Type} & \textbf{Local} & \textbf{MFLOPs/ev\tnote{\colorblue{+}}} & \textbf{Accuracy} \\ 
\toprule
H-First \cite{hfirst}    & Spikes & \no & N/C & 56.1\%          \\
HATS \cite{ncars}        & TS & \no & N/C & 90.2\%          \\
YOLE \cite{yole}         & EH & \yes & 328 & 92.7\%          \\
AsyNet \cite{asynet}     & EH & \yes & 21.5 & 94.4\%          \\
NVS-S \cite{evs}         & Graph & \yes & 5.2 & 91.5\%          \\
EvS-S \cite{evs}         & Graph & \yes & 6.1 & 93.1\%          \\
AEGNN \cite{aegnn}       & Graph & \yes & 0.6\tnote{*} &86.7\%\tnote{*} \\ 
\midrule
\textbf{Ours (software)} & Graph & \yes & 0.07 & 88.0\%          \\
\textbf{Ours (hardware)} & Graph & \yes & 0.07 (INT8) & 87.8\%          \\
\bottomrule
\end{tabular}}

\begin{tablenotes}
    \footnotesize
    \item[+] Algorithmic computational efficiency as the total number of floating-point operations (MFLOPs) per event, at the exception of the EvGNN hardware using INT8 OPs. 1 MAC = 2 OPs.
    \item[*] Based on the AEGNN open-source code on the N-CARS test set and on updated work \cite{dagr_2024}.
    \end{tablenotes}
\end{threeparttable}
\vspace{-0.2cm}
\end{table}

The FPGA resource usage of our \prjname~accelerator is reported in \tab{usage}. The logic usage amounts to 26.4\% of look-up tables (LUTs) as logic, 10.3\% of flip-flops (FFs), and 18.3\% of digital signal processors (DSPs), which are mainly used for MAC units in the graph-convolution module. On-chip memory resource usage amounts to 0.1\% of LUTRAM, 13.2\% of BRAM, and 75.0\% of UltraRAM, which are mainly consumed by the event queues in the graph-building module.

\begin{figure}[!t]
    \centering
    \includegraphics[width=0.95\columnwidth]{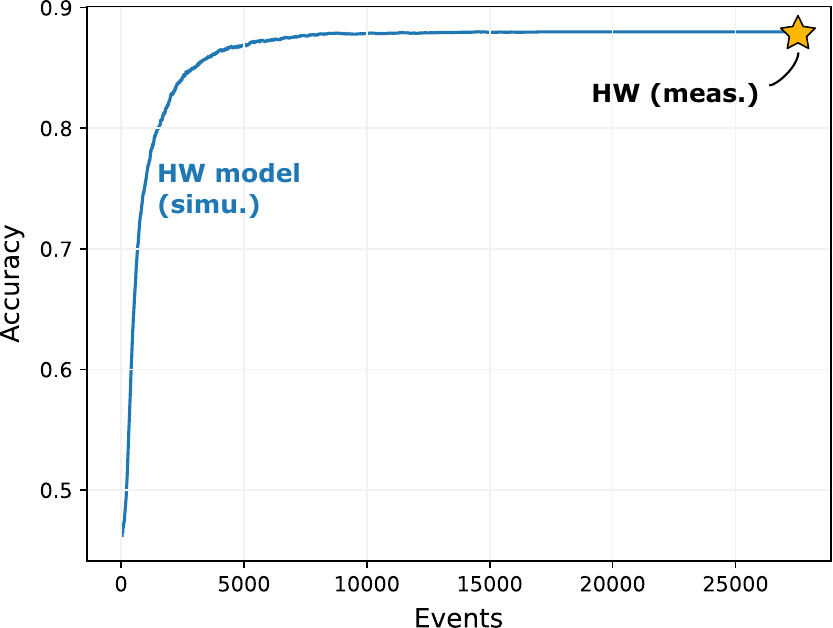}
    \caption{Classification accuracy obtained with \prjname~as a function of the number of events on samples of the N-CARS test set.}
    \label{fig:async_accuracy_full_test}
\end{figure}

State-of-the-art event-driven algorithms are summarized in \tab{accuracy} and compared to \prjname~in terms of computational complexity (MFLOPs per event) and classification accuracy. Methods described as \emph{local} only process a neighboring region around the new event. H-First~\cite{hfirst} uses spiking neural networks (SNNs) to process event-based data, which support event-driven processing but are updated in a global fashion for each new event. HATS~\cite{ncars} exploits time-surface~(TS) representations of event streams, which are processed by global classifiers. YOLE~\cite{yole} and AsyNet~\cite{asynet} use event histograms (EH) to achieve both event-driven and local computation, but require 20$\times$ more parameters than our graph-based approach, degrading the update latency and memory footprint. Finally, the graph-based NVS-S and EvS-S from~\cite{evs}, as well as AEGNN from~\cite{aegnn} are both event-driven and local.
On the one hand, NVS-S and EvS-S adopt a slide graph convolution method to achieve local computation. They first identify nodes and edges to be updated within the K-hop subgraph of the new event, using a sliding window, before applying graph convolution on these elements. On the other hand, AEGNN identifies the increasing neighborhood reached along the GNN layers, processing neighbors from the 1-hop subgraph to the K-hop one.
In contrast to these methods, our approach cuts the computational complexity of convolutional operations by relying on local computation within 1-hop subgraphs, in an event-driven fashion. Despite being optimized for low-footprint edge applications, we achieve classification accuracies of $88.0\%$ (FP32, software) and $87.8\%$ (INT8, hardware) on the N-CARS test set, which outperforms the classification accuracy of AEGNN. The achieved accuracy thus remains competitive with state-of-the-art works, while EvGNN enables an improvement in computational efficiency from two to three orders of magnitude. The dynamic, event-by-event classification accuracy of EvGNN is reported in \fig{async_accuracy_full_test}, which shows that, for low-latency applications, reliable classification can be obtained without having to process the full duration of \mbox{N-CARS} samples, with an average latency per event of 16$\mu$s. Hence, the event-driven GNN hardware acceleration enabled by EvGNN allows exploiting the $\mu$s-level resolution of event-based cameras at the edge.

\subsection{Synchronous vs. Asynchronous GNN Accelerators}
While hardware acceleration for GNNs has started gaining interest~\cite{flowgnn, zhu_mega_2024}, we would like to further emphasize that prior designs all target the \textit{synchronous} processing of large, \textit{static} graphs. Their typical target workloads consist of large-scale databases~\cite{kaggle_2024} or molecular structures~\cite{bongini_molecules_2021}, for which the real-time constraint is typically on the order of milliseconds or more, orders of magnitude above the few-$\mu$s target of the latency-critical event-driven systems considered in this work. Synchronous GNN accelerators can therefore afford a priori analysis of the graph to harness partitioning and execution-rescheduling techniques, which reduce irregular data transfers and thereby improve the energy efficiency~\cite{graphsage,zhu_mega_2024} at the expense of end-to-end latency, from graph loading to inference update. Besides, the synchronous processing in these accelerators requires the storage of the entire graph structure, including all edge-related information. This storage amounts to a $>$50\% memory overhead~\cite{zhu_mega_2024}, which is not desirable for a compact deployment at the edge. These fundamental differences in constraints and use cases preclude a direct performance comparison between both kinds of architectures. Recent partial asynchronous acceleration in~\cite{kryjak_2024} resulted in a latency per event of 4.5ms, which is still beyond the target $\mu$s-level range.

\subsection{Silicon Implementation and Scalability Perspectives}
\begin{figure}[!t]
    \centering
    \hspace*{-0.6cm}  
    \includegraphics[width=0.95\columnwidth]{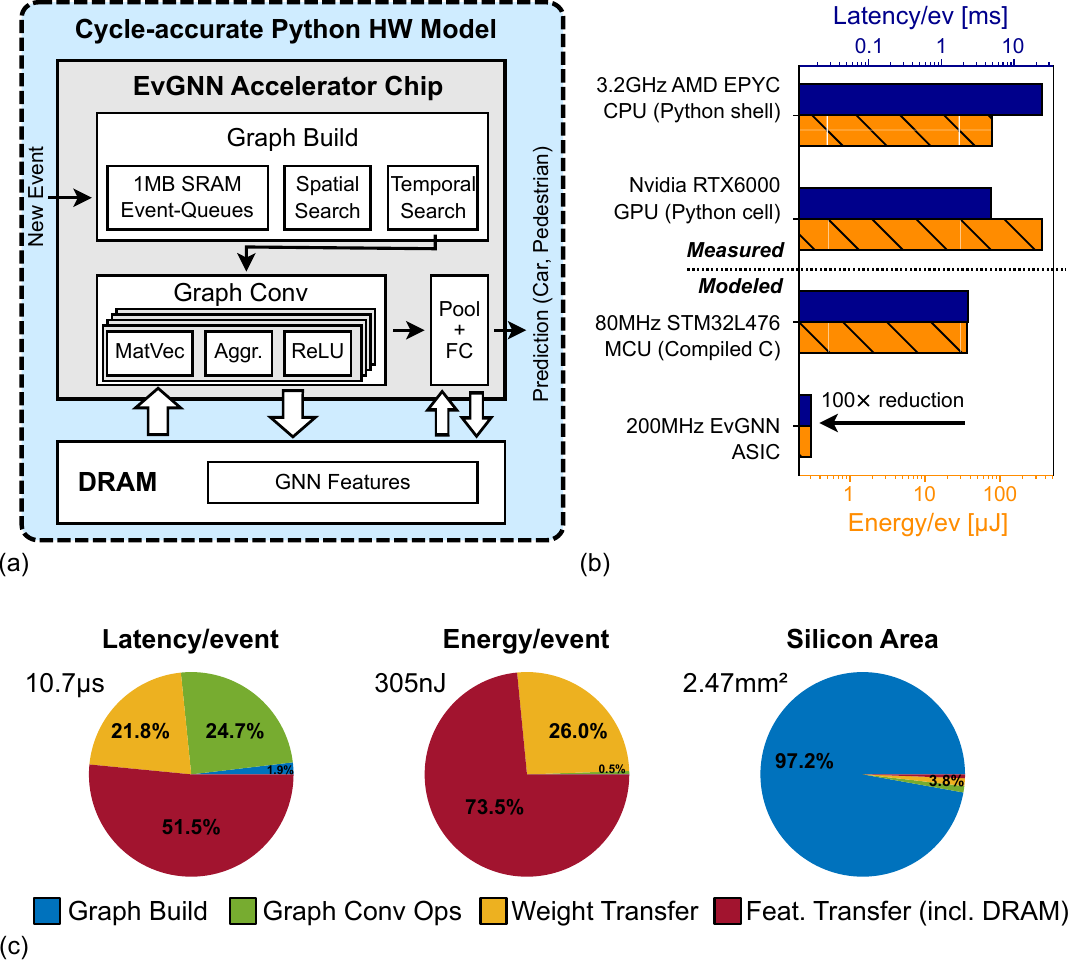}
    \caption{(a) Block diagram of the EvGNN-based system considered for cycle-accurate modeling in a 28nm bulk CMOS process using Python. (b) Per-event latency and energy comparison between EvGNN's algorithm execution on different platforms: a high-performance CPU, a GPU (measured), a commercial MCU, and the EvGNN accelerator (simulated model). (c) Latency, energy and area breakdown of the modeled EvGNN ASIC accelerator. Feature and weight transfers dominate the latency and energy fraction, whereas the event-queue buffer occupies most of the silicon area.}
    \label{fig:silicon_estimates}
\end{figure}

Our FPGA demonstrator paves the way to energy-efficient on-silicon accelerators based on the EvGNN architecture. To estimate the expected latency and energy of such a system, we developed a cycle-accurate model of the accelerator using Python, as described in Fig. \ref{fig:silicon_estimates}(a). We consider a 28nm bulk CMOS process node, a unified core voltage of 0.8V, a frequency of 200MHz, and an off-chip bandwidth for DRAM transfers of 3.2Gb/s. For energy estimates, logic units (MACs) rely on a scaled version of~\cite{horowitz_2014} to 28nm, while both on-chip SRAM buffers and DRAM rely on the CACTI estimator~\cite{cacti_2017}, which is the de-facto tool for modeling memories. The resulting EvGNN application-specific integrated circuit (ASIC) accelerator would occupy a silicon area of 2.5mm$^2$, while achieving a per-event latency and energy of 10.7$\mu s$ and 305nJ on the N-CARS dataset, respectively. In particular, the proposed layer-parallel execution scheme reduces the latency footprint of the graph convolution from 6.9$\mu s$ to 2.6$\mu s$, an improvement by~2.5$\times$.
These numbers are orders of magnitude below what is achievable by high-performance CPUs (AMD EPYC 7763 64-Core Processor 3.2GHz) or GPUs (NVIDIA RTX A6000) running the developed algorithm in Python (Fig. \ref{fig:silicon_estimates}(b)). They are also 100$\times$ better than the estimated metrics for an off-the-shelf deployment on a modeled micro-controller unit (MCU) based on \cite{st_mcu}, where the event-driven algorithm is executed in compiled C code. This gap in performance further emphasizes the importance of building dedicated event-driven GNN accelerators to enable efficient irregular graph processing at the edge.

While our EvGNN accelerator can map event-driven GNNs of any size and achieves excellent latency and energy results on the N-CARS dataset, several challenges may appear when scaling up to larger, real-world scenarios (e.g., \cite{dagr_2024}). In particular, three main challenges would pressure the latency, energy and area breakdown of the accelerator in Fig. \ref{fig:silicon_estimates}(c), for which we identify possible solutions. First, scaling to high-resolution DVS cameras (e.g., QVGA~\cite{dsec_2021} or HD~\cite{mpix_2020} resolutions) would induce a quadratic increase in the size of the event-queue buffer, demanding significant silicon area for an irregular, event-dependent utilization of the storage space. Moving the storage of event queues off-chip would alleviate this area bottleneck at the cost of $\sim 10$\% additional latency and energy, from our model estimates. Avoiding the storage of one event queue per pixel would be another viable option, at the expense of a more complex neighbor-search process. Second, the continuous transfer of data features to/from the DRAM translates into significant latency and energy overheads (Fig.~\ref{fig:silicon_estimates}(c)), which would linearly scale with the layer size. Adding smart caching strategies to efficiently reuse neighbor information between successive events would tackle this overhead. Finally, relying on a 2D array of processing elements with higher data stationarity (e.g., compute-in memory~\cite{kneip_2023}) would improve the latency of every GNN layer execution, which would prove particularly critical for larger layers, as the latency of weight transfers scales quadratically with the layer size. These considerations pave the way for improving the efficiency of future on-silicon GNN accelerator designs.

\section{Conclusion}
\label{sect:conclusion}
We proposed \prjname, the first event-driven GNN accelerator, which enables low-latency, high-accuracy edge inference applications for event-based vision systems. It relies on three key ideas: 
(i) exploiting causality in directed graphs to enable local 1-hop subgraph processing, (ii) a lightweight spatiotemporally decoupled prism neighbor search implemented with flexible event queues, and (iii) a novel layer-parallel execution scheme to reduce the overall processing latency in multi-layer event-driven GNNs. The proposed accelerator was deployed on a Xilinx KV260 MPSoC platform with onboard benchmarking on the \mbox{N-CARS} dataset for car recognition. Our accelerator achieved a prediction accuracy of $87.8\%$ and an average prediction latency of $16 \mu s$ per event, demonstrating the capability to efficiently enable real-time and microsecond-latency event-based vision intelligence at the edge.

\section*{Acknowledgments}
The authors would like to thank Prof. Marian Verhelst (KU Leuven) and Dr. Christoph Posch (Prophesee) for fruitful discussions and feedback.

\printbibliography

@INPROCEEDINGS{navion,
  author={Suleiman, Amr and Zhang, Zhengdong and Carlone, Luca and Karaman, Sertac and Sze, Vivienne},
  booktitle={IEEE Symposium on VLSI Circuits}, 
  title={Navion: A Fully Integrated Energy-Efficient Visual-Inertial Odometry Accelerator for Autonomous Navigation of Nano Drones}, 
  volume={},
  number={},
  pages={133-134},
  year={2018},
  doi={10.1109/VLSIC.2018.8502279}}

@INPROCEEDINGS{1000fps,
  author={Hillebrand, M. and Stevanovic, N. and Hosticka, B.J. and Santos Conde, J.E. and Teuner, A. and Schwarz, M.},
  booktitle={Proceedings of the IEEE Intelligent Vehicles Symposium (Cat. No.00TH8511)}, 
  title={High speed camera system using a CMOS image sensor}, 
  year={2000},
  volume={},
  number={},
  pages={656-661},
  doi={10.1109/IVS.2000.898423}
}

@article{alex,
author = {Krizhevsky, Alex and Sutskever, Ilya and Hinton, Geoffrey E.},
title = {ImageNet classification with deep convolutional neural networks},
year = {2017},
issue_date = {June 2017},
publisher = {Association for Computing Machinery},
address = {New York, NY, USA},
volume = {60},
number = {6},
issn = {0001-0782},
url = {https://doi.org/10.1145/3065386},
doi = {10.1145/3065386},
journal = {Communications of the ACM},
month = {5},
pages = {84–90},
numpages = {7}
}

@inproceedings{resnet,
  title={Deep residual learning for image recognition},
  author={He, Kaiming and Zhang, Xiangyu and Ren, Shaoqing and Sun, Jian},
  booktitle={Proceedings of the IEEE conference on computer vision and pattern recognition},
  pages={770--778},
  year={2016}
}

@INPROCEEDINGS{yolotiny,
  author={Adarsh, Pranav and Rathi, Pratibha and Kumar, Manoj},
  booktitle={International Conference on Advanced Computing and Communication Systems (ICACCS)}, 
  title={YOLO v3-Tiny: Object Detection and Recognition using one stage improved model}, 
  year={2020},
  volume={},
  number={},
  pages={687-694},
  doi={10.1109/ICACCS48705.2020.9074315}
}

@article{sretina,
  title={The silicon retina},
  author={Misha A. Mahowald and Carver Mead},
  journal={Scientific American},
  year={1991},
  volume={264 5},
  pages={
          76-82
        }
}

@ARTICLE{dvs,
  author={Lichtsteiner, Patrick and Posch, Christoph and Delbruck, Tobi},
  journal={IEEE Journal of Solid-State Circuits}, 
  title={A 128$\times$ 128 120 dB 15 $\mu$s Latency Asynchronous Temporal Contrast Vision Sensor}, 
  year={2008},
  volume={43},
  number={2},
  pages={566-576},
  doi={10.1109/JSSC.2007.914337}
}

@inproceedings{time_slice,
    author = {Min Liu and T Delbruck},
    year = {2018},
    title = {Adaptive Time-Slice Block-Matching Optical Flow Algorithm for Dynamic Vision Sensors},
    month = {9},
    booktitle = {British Machine Vision Conference (BMVC)},
    url = {https://doi.org/10.5167/uzh-168589}
}

@inproceedings{evs,
  title={Graph-based asynchronous event processing for rapid object recognition},
  author={Li, Yijin and Zhou, Han and Yang, Bangbang and Zhang, Ye and Cui, Zhaopeng and Bao, Hujun and Zhang, Guofeng},
  booktitle={Proceedings of the IEEE/CVF International Conference on Computer Vision},
  pages={934--943},
  year={2021}
}

@InProceedings{aegnn,
    author    = {Schaefer, Simon and Gehrig, Daniel and Scaramuzza, Davide},
    title     = {AEGNN: Asynchronous Event-Based Graph Neural Networks},
    booktitle = {Proceedings of the IEEE/CVF Conference on Computer Vision and Pattern Recognition (CVPR)},
    month     = {6},
    year      = {2022},
    pages     = {12371-12381}
}

@InProceedings{hugnet,
    author    = {Dalgaty, Thomas and Mesquida, Thomas and Joubert, Damien and Sironi, Amos and Vivet, Pascal and Posch, Christoph},
    title     = {HUGNet: Hemi-Spherical Update Graph Neural Network Applied to Low-Latency Event-Based Optical Flow},
    booktitle = {Proceedings of the IEEE/CVF Conference on Computer Vision and Pattern Recognition (CVPR) Workshops},
    month     = {6},
    year      = {2023},
    pages     = {3952-3961}
}

@ARTICLE{gnn_survey,
  author={Wu, Zonghan and Pan, Shirui and Chen, Fengwen and Long, Guodong and Zhang, Chengqi and Yu, Philip S.},
  journal={IEEE Transactions on Neural Networks and Learning Systems}, 
  title={A Comprehensive Survey on Graph Neural Networks}, 
  year={2021},
  volume={32},
  number={1},
  pages={4-24},
  doi={10.1109/TNNLS.2020.2978386}
}

@article{1stgnn,
  title={Geometric deep learning: Grids, groups, graphs, geodesics, and gauges},
  author={Bronstein, Michael M and Bruna, Joan and Cohen, Taco and Veli{\v{c}}kovi{\'c}, Petar},
  journal={arXiv preprint arXiv:2104.13478},
  year={2021}
}

@article{gcn,
  title={Semi-supervised classification with graph convolutional networks},
  author={Kipf, Thomas N and Welling, Max},
  journal={arXiv preprint arXiv:1609.02907},
  year={2016}
}

@inproceedings{spline,
  title={Splinecnn: Fast geometric deep learning with continuous b-spline kernels},
  author={Fey, Matthias and Lenssen, Jan Eric and Weichert, Frank and M{\"u}ller, Heinrich},
  booktitle={Proceedings of the IEEE conference on computer vision and pattern recognition},
  pages={869--877},
  year={2018}
}

@inproceedings{pointnet,
  title={Pointnet: Deep learning on point sets for 3d classification and segmentation},
  author={Qi, Charles R and Su, Hao and Mo, Kaichun and Guibas, Leonidas J},
  booktitle={Proceedings of the IEEE conference on computer vision and pattern recognition},
  pages={652--660},
  year={2017}
}

@inproceedings{ncars,
  title={HATS: Histograms of averaged time surfaces for robust event-based object classification},
  author={Sironi, Amos and Brambilla, Manuele and Bourdis, Nicolas and Lagorce, Xavier and Benosman, Ryad},
  booktitle={Proceedings of the IEEE conference on computer vision and pattern recognition},
  pages={1731--1740},
  year={2018}
}

@inproceedings{hugnet_ref_13,
  title={End-to-end learning of representations for asynchronous event-based data},
  author={Gehrig, Daniel and Loquercio, Antonio and Derpanis, Konstantinos G and Scaramuzza, Davide},
  booktitle={Proceedings of the IEEE/CVF International Conference on Computer Vision},
  pages={5633--5643},
  year={2019}
}

@inproceedings{flowgnn,
  title={FlowGNN: A Dataflow Architecture for Real-Time Workload-Agnostic Graph Neural Network Inference},
  author={Sarkar, Rishov and Abi-Karam, Stefan and He, Yuqi and Sathidevi, Lakshmi and Hao, Cong},
  booktitle={IEEE International Symposium on High-Performance Computer Architecture (HPCA)},
  pages={1099--1112},
  year={2023}
}

@inproceedings{blockgnn,
  title={Blockgnn: Towards efficient gnn acceleration using block-circulant weight matrices},
  author={Zhou, Zhe and Shi, Bizhao and Zhang, Zhe and Guan, Yijin and Sun, Guangyu and Luo, Guojie},
  booktitle={ACM/IEEE Design Automation Conference (DAC)},
  pages={1009--1014},
  year={2021}
}

@inproceedings{eventq,
  title={Learning an event sequence embedding for dense event-based deep stereo},
  author={Tulyakov, Stepan and Fleuret, Francois and Kiefel, Martin and Gehler, Peter and Hirsch, Michael},
  booktitle={Proceedings of the IEEE/CVF International Conference on Computer Vision},
  pages={1527--1537},
  year={2019}
}

@article{quant1,
  title={Quantizing deep convolutional networks for efficient inference: A whitepaper},
  author={Krishnamoorthi, Raghuraman},
  journal={arXiv preprint arXiv:1806.08342},
  year={2018}
}

@inproceedings{quant3,
  title={Quantization and training of neural networks for efficient integer-arithmetic-only inference},
  author={Jacob, Benoit and Kligys, Skirmantas and Chen, Bo and Zhu, Menglong and Tang, Matthew and Howard, Andrew and Adam, Hartwig and Kalenichenko, Dmitry},
  booktitle={Proceedings of the IEEE conference on computer vision and pattern recognition},
  pages={2704--2713},
  year={2018}
}

@article{pyg,
  title={Fast graph representation learning with PyTorch Geometric},
  author={Fey, Matthias and Lenssen, Jan Eric},
  journal={arXiv preprint arXiv:1903.02428},
  year={2019}
}

@InProceedings{dense_frame,
author = {Nguyen, Anh and Do, Thanh-Toan and Caldwell, Darwin G. and Tsagarakis, Nikos G.},
title = {Real-Time 6DOF Pose Relocalization for Event Cameras With Stacked Spatial LSTM Networks},
booktitle = {Proceedings of the IEEE/CVF Conference on Computer Vision and Pattern Recognition (CVPR) Workshops},
month = {6},
year = {2019}
}

@InProceedings{asynet,
author="Messikommer, Nico
and Gehrig, Daniel
and Loquercio, Antonio
and Scaramuzza, Davide",
editor="Vedaldi, Andrea
and Bischof, Horst
and Brox, Thomas
and Frahm, Jan-Michael",
title="Event-Based Asynchronous Sparse Convolutional Networks",
booktitle="Computer Vision -- ECCV 2020",
year="2020",
pages="415--431",
isbn="978-3-030-58598-3"
}

@article{event_survey,
  title={Event-based vision: A survey},
  author={Gallego, Guillermo and Delbr{\"u}ck, Tobi and Orchard, Garrick and Bartolozzi, Chiara and Taba, Brian and Censi, Andrea and Leutenegger, Stefan and Davison, Andrew J and Conradt, J{\"o}rg and Daniilidis, Kostas and others},
  journal={IEEE transactions on pattern analysis and machine intelligence},
  volume={44},
  number={1},
  pages={154--180},
  year={2020},
  publisher={IEEE}
}

@inproceedings{diffpool,
 author = {Ying, Zhitao and You, Jiaxuan and Morris, Christopher and Ren, Xiang and Hamilton, Will and Leskovec, Jure},
 booktitle = {Advances in Neural Information Processing Systems},
 pages = {},
 title = {Hierarchical Graph Representation Learning with Differentiable Pooling},
 volume = {31},
 year = {2018}
}

@inproceedings{eigenpool,
  title={Graph convolutional networks with eigenpooling},
  author={Ma, Yao and Wang, Suhang and Aggarwal, Charu C and Tang, Jiliang},
  booktitle={Proceedings of the 25th ACM SIGKDD international conference on knowledge discovery \& data mining},
  pages={723--731},
  year={2019}
}

@inproceedings{gnn_receptive_1,
  title={A brief review of receptive fields in graph convolutional networks},
  author={Quan, Pei and Shi, Yong and Lei, Minglong and Leng, Jiaxu and Zhang, Tianlin and Niu, Lingfeng},
  booktitle={IEEE/WIC/ACM International Conference on Web Intelligence-Companion Volume},
  pages={106--110},
  year={2019}
}

@article{gnn_receptive_2,
title = {Multireceptive field: An adaptive path aggregation graph neural framework for hyperspectral image classification},
journal = {Expert Systems with Applications},
volume = {217},
pages = {119508},
year = {2023},
issn = {0957-4174},
doi = {https://doi.org/10.1016/j.eswa.2023.119508},
url = {https://www.sciencedirect.com/science/article/pii/S095741742300009X},
author = {Zhili Zhang and Yao Ding and Xiaofeng Zhao and L. Siye and Nengjun Yang and Yaoming Cai and Ying Zhan}
}

@article{hfirst,
  title={HFirst: A temporal approach to object recognition},
  author={Orchard, Garrick and Meyer, Cedric and Etienne-Cummings, Ralph and Posch, Christoph and Thakor, Nitish and Benosman, Ryad},
  journal={IEEE transactions on pattern analysis and machine intelligence},
  volume={37},
  number={10},
  pages={2028--2040},
  year={2015},
  publisher={IEEE}
}

@InProceedings{yole,
    author = {Cannici, Marco and Ciccone, Marco and Romanoni, Andrea and Matteucci, Matteo},
    title = {Asynchronous Convolutional Networks for Object Detection in Neuromorphic Cameras},
    booktitle = {Proceedings of the IEEE/CVF Conference on Computer Vision and Pattern Recognition (CVPR) Workshops},
    month = {6},
    year = {2019}
}

@inproceedings{edgeeye,
  title={Edgeeye: A long-range energy-efficient vision node for long-term edge computing},
  author={Benninger, Simon and Magno, Michele and Gomez, Andres and Benini, Luca},
  booktitle={International Green and Sustainable Computing Conference (IGSC)},
  pages={1--8},
  year={2019}
}

@INPROCEEDINGS{tinytracker,
  author={Bonazzi, Pietro and Rüegg, Thomas and Bian, Sizhen and Li, Yawei and Magno, Michele},
  booktitle={IEEE SENSORS}, 
  title={TinyTracker: Ultra-Fast and Ultra-Low-Power Edge Vision In-Sensor for Gaze Estimation}, 
  year={2023},
  volume={},
  number={},
  pages={1-4},
  doi={10.1109/SENSORS56945.2023.10325167}}

@article{pulp,
  title={PULP: A ultra-low power parallel accelerator for energy-efficient and flexible embedded vision},
  author={Conti, Francesco and Rossi, Davide and Pullini, Antonio and Loi, Igor and Benini, Luca},
  journal={Journal of Signal Processing Systems},
  volume={84},
  pages={339--354},
  year={2016},
  publisher={Springer}
}

@article{sparseyolo,
  title={Sparse-YOLO: Hardware/software co-design of an FPGA accelerator for YOLOv2},
  author={Wang, Zixiao and Xu, Ke and Wu, Shuaixiao and Liu, Li and Liu, Lingzhi and Wang, Dong},
  journal={IEEE Access},
  volume={8},
  pages={116569--116585},
  year={2020},
  publisher={IEEE}
}

@ARTICLE{flyingiot,
  author={Genc, Hasan and Zu, Yazhou and Chin, Ting-Wu and Halpern, Matthew and Reddi, Vijay Janapa},
  journal={IEEE Micro}, 
  title={Flying IoT: Toward Low-Power Vision in the Sky}, 
  year={2017},
  volume={37},
  number={6},
  pages={40-51},
  doi={10.1109/MM.2017.4241339}}

@ARTICLE{lopecs,
  author={Tang, Jie and Liu, Shaoshan and Liu, Liangkai and Yu, Bo and Shi, Weisong},
  journal={IEEE Access}, 
  title={LoPECS: A Low-Power Edge Computing System for Real-Time Autonomous Driving Services}, 
  year={2020},
  volume={8},
  number={},
  pages={30467-30479},
  doi={10.1109/ACCESS.2020.2970728}}

@ARTICLE{robot_sense_1,
  author={Huang, Le and Wu, Gongping and Tang, Wenjie and Wu, Yi},
  journal={IEEE Access}, 
  title={Obstacle Distance Measurement Under Varying Illumination Conditions Based on Monocular Vision Using a Cable Inspection Robot}, 
  year={2021},
  volume={9},
  number={},
  pages={55955-55973},
  doi={10.1109/ACCESS.2021.3070877}}

@inproceedings{mobilenetv3,
  title={Searching for mobilenetv3},
  author={Howard, Andrew and Sandler, Mark and Chu, Grace and Chen, Liang-Chieh and Chen, Bo and Tan, Mingxing and Wang, Weijun and Zhu, Yukun and Pang, Ruoming and Vasudevan, Vijay and others},
  booktitle={Proceedings of the IEEE/CVF international conference on computer vision},
  pages={1314--1324},
  year={2019}
}

@ARTICLE{dualconv,
  author={Zhong, Jiachen and Chen, Junying and Mian, Ajmal},
  journal={IEEE Transactions on Neural Networks and Learning Systems}, 
  title={DualConv: Dual Convolutional Kernels for Lightweight Deep Neural Networks}, 
  year={2023},
  volume={34},
  number={11},
  pages={9528-9535},
  doi={10.1109/TNNLS.2022.3151138}}

@ARTICLE{cnn_survey,
  author={Li, Zewen and Liu, Fan and Yang, Wenjie and Peng, Shouheng and Zhou, Jun},
  journal={IEEE Transactions on Neural Networks and Learning Systems}, 
  title={A Survey of Convolutional Neural Networks: Analysis, Applications, and Prospects}, 
  year={2022},
  volume={33},
  number={12},
  pages={6999-7019},
  doi={10.1109/TNNLS.2021.3084827}}

@article{hop_define,
  title={An efficient parallel algorithm of n-hop neighborhoods on graphs in distributed environment},
  author={Liu, Wenjie and Li, Zhanhuai},
  journal={Frontiers of Computer Science},
  volume={13},
  pages={1309--1325},
  year={2019},
  publisher={Springer}
}

@InProceedings{static_event_graph_1,
author = {Mitrokhin, Anton and Hua, Zhiyuan and Fermuller, Cornelia and Aloimonos, Yiannis},
title = {Learning Visual Motion Segmentation Using Event Surfaces},
booktitle = {Proceedings of the IEEE/CVF Conference on Computer Vision and Pattern Recognition (CVPR)},
pages={14414--14423},
month = {6},
year = {2020}
}

@ARTICLE{static_event_graph_2,
  author={Bi, Yin and Chadha, Aaron and Abbas, Alhabib and Bourtsoulatze, Eirina and Andreopoulos, Yiannis},
  journal={IEEE Transactions on Image Processing}, 
  title={Graph-Based Spatio-Temporal Feature Learning for Neuromorphic Vision Sensing}, 
  year={2020},
  volume={29},
  number={},
  pages={9084-9098},
  doi={10.1109/TIP.2020.3023597}}

@article{lp,
  title={General considerations on the use of the weighted lp norm as an empirical distance measure},
  author={Brimberg, Jack and Love, Robert F},
  journal={Transportation Science},
  volume={27},
  number={4},
  pages={341--349},
  year={1993},
  publisher={INFORMS}
}

@inproceedings{agcn,
  title={Adaptive graph convolutional neural networks},
  author={Li, Ruoyu and Wang, Sheng and Zhu, Feiyun and Huang, Junzhou},
  booktitle={Proceedings of the AAAI conference on artificial intelligence},
  volume={32},
  number={1},
  year={2018}
}

@inproceedings{dcnn,
 author = {Atwood, James and Towsley, Don},
 booktitle = {Advances in Neural Information Processing Systems},
 pages = {},
 title = {Diffusion-Convolutional Neural Networks},
 volume = {29},
 year = {2016}
}

@article{rinaldo_bootstrap,
author = {Alessandro Rinaldo and Larry Wasserman and Max G’Sell},
title = {{Bootstrapping and sample splitting for high-dimensional, assumption-lean inference}},
volume = {47},
journal = {The Annals of Statistics},
number = {6},
publisher = {Institute of Mathematical Statistics},
pages = {3438 -- 3469},
year = {2019},
doi = {10.1214/18-AOS1784},
URL = {https://doi.org/10.1214/18-AOS1784}
}

@inproceedings{zhu_mega_2024,
    author={Zhu, Zeyu and Li, Fanrong and Li, Gang and Liu, Zejian and Mo, Zitao and Hu, Qinghao and Liang, Xiaoyao and Cheng, Jian},
    booktitle={2024 IEEE International Symposium on High-Performance Computer Architecture (HPCA)}, 
    title={{MEGA: A Memory-Efficient GNN Accelerator Exploiting Degree-Aware Mixed-Precision Quantization}}, 
    year={2024},
    volume={},
    number={},
    pages={124-138},
    doi={10.1109/HPCA57654.2024.00020},
    url={https://ieeexplore.ieee.org/abstract/document/10476431}}

@online{kaggle_2024,
    author = {Kaggle},
    title = {{Journal Ranking Databases}},
    year = 2024,
    url = {https://www.kaggle.com/datasets/xabirhasan/journal-
    ranking-dataset}
}

@article{bongini_molecules_2021,
author = {Pietro Bongini and Monica Bianchini and Franco Scarselli},
title = {{Molecular generative Graph Neural Networks for Drug Discovery}},
journal = {Neurocomputing},
volume = {450},
pages = {242-252},
year = {2021},
doi = {https://doi.org/10.1016/j.neucom.2021.04.039},
url = {https://www.sciencedirect.com/science/article/pii/S0925231221005737},
}

@article{dagr_2024,
author={Gehrig, D. and Scaramuzza, D.},
title = {{Low-latency automotive vision with event cameras}},
journal = {Nature},
volume = {629},
pages = {1034-1040},
year = {2024},
doi = {https://doi.org/10.1038/s41586-024-07409-w}
}

@article{dsec_2021,
  author={Gehrig, Mathias and Aarents, Willem and Gehrig, Daniel and Scaramuzza, Davide},
  title={{DSEC: A Stereo Event Camera Dataset for Driving Scenarios}}, 
  journal={{IEEE Robotics and Automation Letters}}, 
  volume={6},
  number={3},
  pages={4947-4954},
  year={2021},
  doi={10.1109/LRA.2021.3068942}
}

@inproceedings{mpix_2020,
 author = {Perot, Etienne and de Tournemire, Pierre and Nitti, Davide and Masci, Jonathan and Sironi, Amos},
 booktitle = {Advances in Neural Information Processing Systems},
 editor = {H. Larochelle and M. Ranzato and R. Hadsell and M.F. Balcan and H. Lin},
 pages = {16639--16652},
 publisher = {Curran Associates, Inc.},
 title = {{Learning to Detect Objects with a 1 Megapixel Event Camera}},
 url = {https://proceedings.neurips.cc/paper_files/paper/2020/file/c213877427b46fa96cff6c39e837ccee-Paper.pdf},
 volume = {33},
 year = {2020}
}

@online{st_mcu,
    author = {ST Microelectronics},
    title = {{STM32L476xx: Ultra-low-power Arm ® Cortex ® -M4 32-bit MCU+FPU, 100DMIPS,up to 1MB Flash, 128 KB SRAM, USB OTG FS, LCD, ext. SMPS}},
    url = {https://docs.rs-online.com/d8ed/A700000008303040.pdf},
    year = {2019}
}

@inproceedings{horowitz_2014,
  author={Horowitz, Mark},
  booktitle={2014 IEEE International Solid-State Circuits Conference Digest of Technical Papers (ISSCC)}, 
  title={{1.1 Computing's energy problem (and what we can do about it)}}, 
  year={2014},
  volume={},
  number={},
  pages={10-14},
  DOI={10.1109/ISSCC.2014.6757323}
}

@article{cacti_2017,
author = {Balasubramonian, Rajeev and Kahng, Andrew B. and Muralimanohar, Naveen and Shafiee, Ali and Srinivas, Vaishnav},
title = {{CACTI 7: New Tools for Interconnect Exploration in Innovative Off-Chip Memories}},
journal = {{ACM Trans. Archit. Code Optim.}},
volume = {14},
number = {2},
url = {https://doi.org/10.1145/3085572},
DOI = {10.1145/3085572},
year = {2017},
pages = {1-25}
}

@inproceedings{int8_edp,
  title={Performance-efficiency trade-off of low-precision numerical formats in deep neural networks},
  author={Carmichael, Zachariah and Langroudi, Hamed F and Khazanov, Char and Lillie, Jeffrey and Gustafson, John L and Kudithipudi, Dhireesha},
  booktitle={Proceedings of the conference for next generation arithmetic 2019},
  pages={1--9},
  year={2019}
}

@inproceedings{deep_positron,
  title={Deep positron: A deep neural network using the posit number system},
  author={Carmichael, Zachariah and Langroudi, Hamed F and Khazanov, Char and Lillie, Jeffrey and Gustafson, John L and Kudithipudi, Dhireesha},
  booktitle={2019 Design, Automation \& Test in Europe Conference \& Exhibition (DATE)},
  pages={1421--1426},
  year={2019},
  organization={IEEE}
}

@article{arafat_drones_2023,
  author={Arafat, M.Y. and Alam, M.M. and Moh, S.},
  title={{Vision-Based Navigation Techniques for Unmanned Aerial Vehicles: Review and Challenges}}, 
  journal={{Drones}}, 
  volume={7},
  number={89},
  pages={1-41},
  year={2023},
  doi={10.3390/drones7020089 }
}

@article{gupta_cars_2021,
  author={Gupta, A. and Anpalagan, A. and Guan, L. and Khwaja, A.S.},
  title={{Deep learning for object detection and scene perception in self-driving cars: Survey, challenges, and open issues}}, 
  journal={{Array}}, 
  volume={10},
  pages={1-20},
  year={2021},
  doi={10.1016/j.array.2021.100057}
}

@article{graphsage,
  title={{Inductive Representation Learning on Large Graphs}},
  author={Hamilton, William L. and Ying, Rex and Leskovec, Jure},
  journal={arXiv preprint arXiv:1706.02216v4},
  year={2018}
}

@ARTICLE{kneip_2023,
  author={Kneip, Adrian and Lefebvre, Martin and Verecken, Julien and Bol, David},
  journal={{IEEE Journal of Solid-State Circuits}}, 
  title={{IMPACT: A 1-to-4b 813-TOPS/W 22-nm FD-SOI Compute-in-Memory CNN Accelerator Featuring a 4.2-POPS/W 146-TOPS/mm2 CIM-SRAM With Multi-Bit Analog Batch-Normalization}}, 
  year={2023},
  volume={58},
  number={7},
  pages={1871-1884},
  doi={10.1109/JSSC.2023.3269098}
}

@article{kryjak_2024,
  title={{Embedded Graph Convolutional Networks for Real-Time Event Data Processing on SoC FPGAs}},
  author={Jeziorek, Kamil and Wzorek, Piotr and Blachut, Krzysztof and Pinna, Andrea and Kryjak, Tomasz},
  journal={arXiv preprint arXiv:2406.07318},
  pages={1-20},
  year={2024}
}



\vspace{-1.0cm}

\begin{IEEEbiography}[{\includegraphics[width=1in,height=1.25in,clip,keepaspectratio]{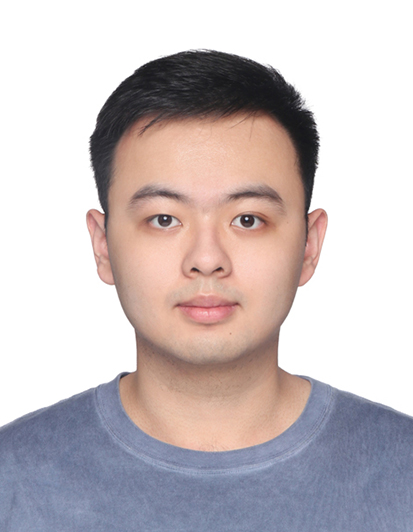}}]{Yufeng Yang} received his bachelor's degree in Electronic and Information Engineering from Beihang University in 2020, and the M.Sc. degree in Electrical Engineering from Delft University of Technology (TU Delft) in 2023 under the supervision of Prof. Charlotte Frenkel. His research focuses on the hardware-algorithm co-design of modern neural networks and on the design of efficient AI accelerators for resource-constrained embedded hardware platforms.
\end{IEEEbiography}

\vspace{-1.0cm}

\begin{IEEEbiography}[{\includegraphics[width=1in,height=1.25in,clip,keepaspectratio]{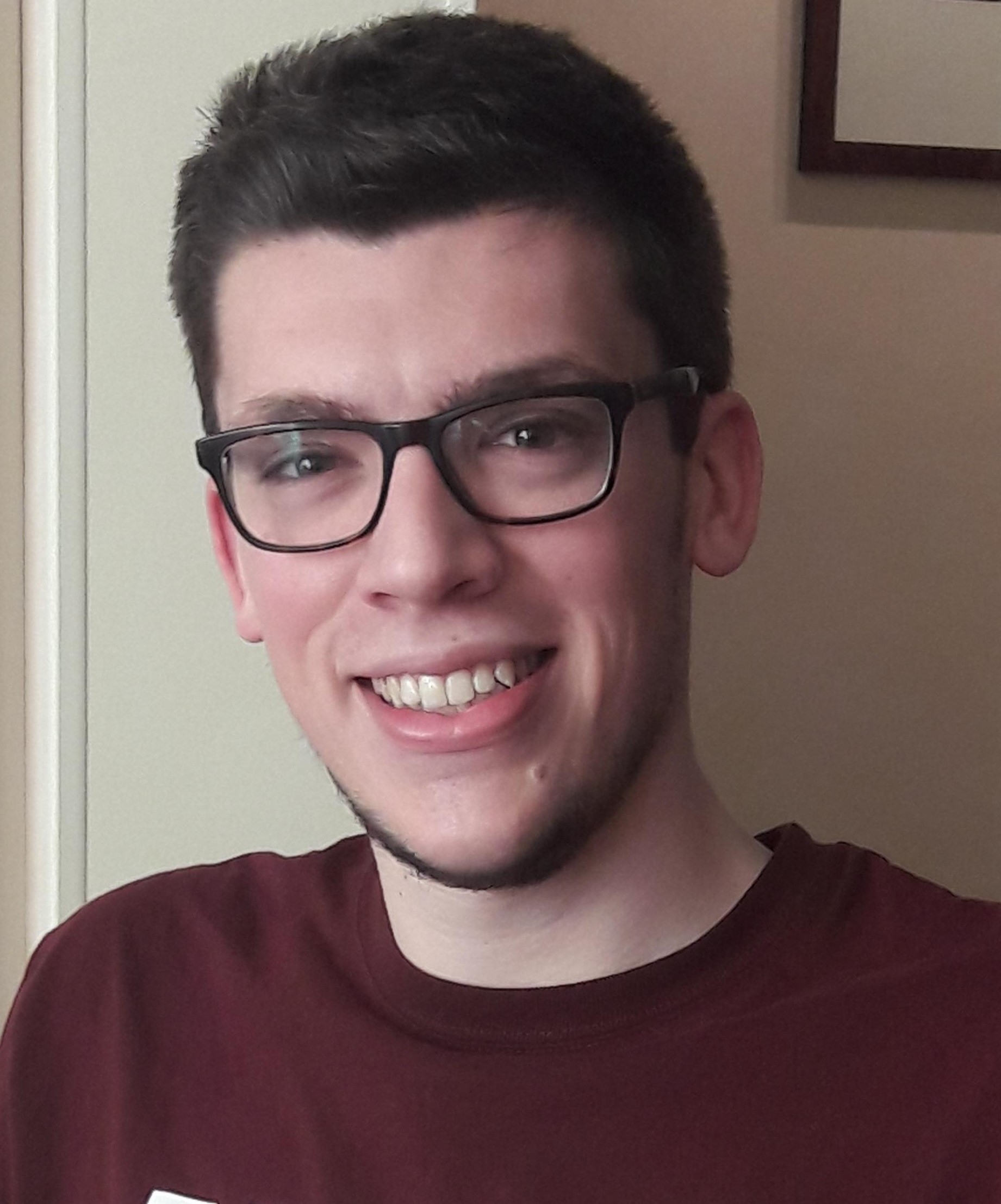}}]{Adrian Kneip}(Member, IEEE) received the M.Sc. degree (\textit{summa cum laude}) in Electrical Engineering from the Universit\'e catholique de Louvain (UCLouvain), Louvain-la-Neuve, Belgium, in 2019. He then joined UCLouvain's Electronics Circuits and Systems (ECS) group, where he obtained the Ph.D. degree in 2024 under the supervision of Prof. D. Bol. His research interests notably include the design of ultra-low-power digital ICs, as well as analog/mixed-signal accelerators for edge-AI chips. He has a particular interest for SRAM-based in-memory computing and hardware/software co-design aspects. Recently, he has also shown a rising interest event-based computing techniques, with a focus on event-driven graph neural networks and their hardware acceleration. Dr. Kneip is the author or co-author of several research papers in IEEE conferences and journals, receiving the Best Student Paper Award for the 2022's ESSCIRC-ESSDERC conference. He also serves as reviewer for various IEEE SSCS and CAS journals, wherein IEEE ESSCERC, ISCAS, JSSC and TCAS-I/-II. He also was Student Representative to the IEEE Benelux Section from 2020 to 2023.
\end{IEEEbiography}

\vfill
\newpage

\begin{IEEEbiography}[{\includegraphics[width=1.25in,height=1.3in,clip,keepaspectratio]{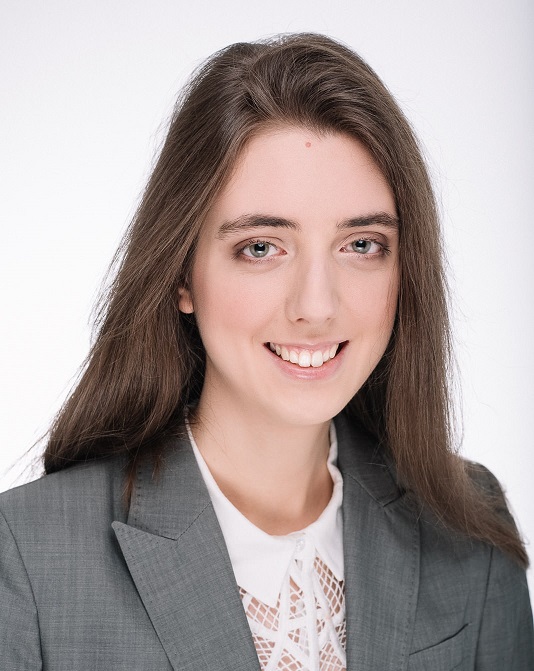}}]{Charlotte Frenkel}
(Member, IEEE) received the M.Sc. degree (\textit{summa cum laude}) in Electromechanical Engineering and the Ph.D. degree in Engineering Science from Universit\'e catholique de Louvain (UCLouvain), Louvain-la-Neuve, Belgium in 2015 and 2020, respectively. In February 2020, she joined the Institute of Neuroinformatics, UZH and ETH Zurich, Switzerland, as a postdoctoral researcher. She is an Assistant Professor at Delft University of Technology, Delft, The Netherlands, since July 2022, and holds a Visiting Faculty Researcher position with Google since October 2024.

Her research aims at bridging the bottom-up (bio-inspired) and top-down (engineering-driven) design approaches toward neuromorphic intelligence, with a focus on hardware-algorithm co-design for (Neuro)AI, digital hardware accelerators, and brain-inspired on-device learning.

Dr. Frenkel received a best paper award at the IEEE International Symposium on Circuits and Systems (ISCAS) 2020 conference in the \textit{Neural Networks} track, and her Ph.D. thesis was awarded the FNRS-FWO / Nokia Bell Scientific Award 2021 and the FNRS-FWO / IBM Innovation Award 2021. In 2023, she was awarded prestigious Veni and AiNed Fellowship grants from the Dutch Research Council (NWO). She presented several invited talks, including keynotes at the tinyML EMEA technical forum 2021 and at the Neuro-Inspired Computational Elements (NICE) neuromorphic conference 2021. She serves or has served as a program co-chair of NICE 2023-2024 and of the tinyML Research Symposium 2024, as a co-lead of the NeuroBench initiative for benchmarks in neuromorphic computing since 2022, as a TPC member of IEEE ESSERC for 2022-2024, and as an associate editor for the IEEE Transactions on Biomedical Circuits and Systems since 2022.
\end{IEEEbiography}

\vfill

\end{document}